\documentclass[journal]{IEEEtran}

\usepackage{mathtools}
\usepackage{array}
\usepackage{float}
\usepackage{hyperref}
\usepackage[pdftex]{graphicx}
\graphicspath{{./graphics/}}
\usepackage{subfigure}
\usepackage{multicol}
\usepackage{multirow}
\usepackage{paralist}
\usepackage{bm}
\usepackage{amsmath,amsthm,amssymb,amsfonts}
\usepackage[usenames,dvipsnames]{color}
\usepackage{dsfont}
\usepackage{url}
\usepackage{rotating}
\usepackage{enumerate}
\usepackage{color}
\usepackage{slashbox}

\usepackage{ifpdf}
\usepackage{cite}

\usepackage{textcomp,booktabs}
\usepackage{colortbl}
\usepackage{gensymb}

\makeatletter
\newif\if@restonecol
\makeatother

\usepackage[linesnumbered,ruled]{algorithm2e}
\usepackage{algpseudocode}
\newcommand{\jb}{\mathbf{J}}
\newcommand{\mb}{\mathbf{M}}
\newcommand{\cb}{\mathbf{C}}
\newcommand{\db}{\mathbf{D}}
\newcommand{\gb}{\mathbf{G}}

\newcommand{\s}{\mathbf{s}}
\newcommand{\ab}{\mathbf{a}}
\newcommand{\eb}{\mathbf{e}}
\newcommand{\et}{\mathbf{\eta}}
\newcommand{\ph}{\mathbf{\phi}}
\newcommand{\ta}{\mathbf{\tau}}
\newcommand{\argmax}{\operatornamewithlimits{argmax}}
\newcommand{\argmin}{\operatornamewithlimits{argmin}}

\begin{document}
%
%

\title{Multi Pseudo Q-learning Based Deterministic Policy Gradient for Tracking Control of Autonomous Underwater Vehicles}

\author{Wenjie~Shi,
        Shiji~Song,~\IEEEmembership{Senior~Member,~IEEE},
        Cheng~Wu
        and~C.~L.~Philip~Chen,~\IEEEmembership{Fellow,~IEEE}

\thanks{This work was supported by the National Natural Science Foundation of China under Grant 41427806.}

\thanks{Wenjie~Shi, Shiji~Song, Cheng~Wu are with the Department of Automation, Tsinghua University, Beijing 100084, China (email: shiwj16@mails.tsinghua.edu.cn, shijis@mail.tsinghua.edu.cn, wuc@tsinghua.edu.cn.)}
\thanks{C. L. Philip Chen is with Department of Computer and Information Science, Faculty of Science and Technology, University of Macau, Macau 99999, China (e-mail: philipchen@umac.mo)}
}

\maketitle

\begin{abstract}
 This paper investigates trajectory tracking problem for a class of underactuated autonomous underwater vehicles (AUVs) with unknown dynamics and constrained inputs. Different from existing policy gradient methods which employ single actor-critic but cannot realize satisfactory tracking control accuracy and stable learning, our proposed algorithm can achieve high-level tracking control accuracy of AUVs and stable learning by applying a hybrid actors-critics architecture, where multiple actors and critics are trained to learn a deterministic policy and action-value function, respectively. Specifically, for the critics, the expected absolute Bellman error based updating rule is used to choose the worst critic to be updated in each time step. Subsequently, to calculate the loss function with more accurate target value for the chosen critic, Pseudo Q-learning, which uses sub-greedy policy to replace the greedy policy in Q-learning, is developed for continuous action spaces, and Multi Pseudo Q-learning (MPQ) is proposed to reduce the overestimation of action-value function and to stabilize the learning. As for the actors, deterministic policy gradient is applied to update the weights, and the final learned policy is defined as the average of all actors to avoid large but bad updates. Moreover, the stability analysis of the learning is given qualitatively. The effectiveness and generality of the proposed MPQ-based Deterministic Policy Gradient (MPQ-DPG) algorithm are verified by the application on AUV with two different reference trajectories. And the results demonstrate high-level tracking control accuracy and stable learning of MPQ-DPG. Besides, the results also validate that increasing the number of the actors and critics will further improve the performance.
\end{abstract}


\IEEEpeerreviewmaketitle

\section{Introduction}
 \IEEEPARstart{A}{utonomous} underwater vehicles (AUVs) have become increasingly attractive and been widely studied for various mission scenarios of interest such as oceanic resources exploration, seafloor survey and oceanographic mapping, where it is dangerous for human to operate in underwater environment without the help of AUVs. Hence, it is meaningful to develop intelligent control methods for AUVs to move in an unmanned way. And two recent survey papers \cite{Singh2011Control,das2016cooperative} are recommended for the latest developments in the design and motion control of AUVs. However, many offshore applications of AUVs, such as trajectory tracking problem, are very challenging due to three main difficulties. Firstly, AUVs are highly nonlinear multi-input multi-output systems with strong coupling and time-varying hydrodynamic coefficients of dynamics. Secondly, AUVs as well as environment models are often poorly known. Thirdly, most AUVs are designed as underactuated, that is, their degrees of freedom (DOF) are greater than the number of independent actuators. All of these make it necessary to further study trajectory tracking problem of AUVs.

 Over the past decades, a large number of control methods to overcome these difficulties have been proposed by researchers. There exist different motion controllers for trajectory tracking \cite{refsnes2008model,li2016trajectory}, way-point tracking \cite{kim2015tracking}, path planning \cite{zhu2013dynamic,cui2016mutual} and path-following \cite{peng2017output,xiang2016path} in the literature. But in studies about trajectory tracking problem of AUVs, the dynamic models are often decoupled or linearized to enable potential applications of various classic controllers \cite{refsnes2008model,healey1993multivariable,Subudhi2013A, He2015Vibration}. For example, an output feedback controller has been derived with two decoupled plant models in \cite{refsnes2008model}. A state feedback controller has been designed for tracking of AUVs in \cite{healey1993multivariable}, where the model was linearized with a constant forward velocity and decoupled into three separate systems. And in \cite{Subudhi2013A,He2015Vibration}, the controllers were based on the linearized AUV models. Unfortunately, these mentioned works cannot address all three issues.

 Due to the limitations of the above methods and the amazing self-learning ability of intelligent control, researchers have shown great interest in developing learning-based control methods such as adaptive neural network control \cite{cui2010leader,wang2015neural,he2016adaptive1,he2016adaptive2,sun2017adaptive,wei2017adaptive} and reinforcement learning (RL) control \cite{Konar2013ADI,song2015multiple,kamalapurkar2017model,mu2017air, ouyang2017reinforcement}. And adaptive neural network control has been successfully applied to multi-link robots \cite{he2016adaptive1, he2016adaptive2}, biped robots \cite{sun2017adaptive} and marine vessels \cite{wei2017adaptive}. For example, both full state feedback control and output feedback control have been proposed for trajectory tracking of a marine surface full-actuated vessel in \cite{wei2017adaptive}, where radial basis function neural network (RBF-NN) was used to approximate unknown model parameters of the vessel. But as mentioned above, AUV model considered in this paper is underactuated, highly nonlinear and strongly coupled, which make general NNs or RBF-NNs incapable to effectively approximate unknown model due to their limited approximation ability.

 Different from adaptive neural network control, the unknown model does not need to be approximated in RL control methods, which realize performance optimization of a system by ongoing interacting with environment. Specifically, the controller is described as a policy, whose performance is evaluated by action-value function, and Bellman optimality equation is applied to recursively solve action-value function. So far, some RL control methods have been studied to design applicable controller for motion control of AUVs \cite{gaskett1999reinforcement, carreras2005behavior,el2013two,de2015trajectory}. For example, neural Q-learning and direct policy search have been proposed for tracking control of AUVs in \cite{carreras2005behavior,el2013two}, but these methods are only feasible for concrete action spaces. More recent is the work done in \cite{de2015trajectory} where a gaussian process was used to model the policy for continuous action space, but its tracking control accuracy cannot be guaranteed.

 Recently, with the development of some new techniques such as batch learning, experience replay and batch normalization in training deep neural networks (DNNs), deep reinforcement learning (DRL) has shown powerful capabilities to solve complex tasks without much prior knowledge, such as robotic motion control \cite{tai2017virtual}, longitudinal control of ALVs \cite{huang2017parameterized}, control of a quadrotor \cite{hwangbo2017control,mannucci2017safe} and autonomous driving \cite{sallab2017deep}. And in \cite{mnih2015human}, ``Deep Q Network" (DQN) was proposed to realize human-level control for many challenging tasks. However, DQN cannot handle continuous action spaces for problems with high-dimensional state spaces. Consequently, it is impossible for DQN to directly scale to trajectory tracking problem of AUVs, because both state and action of AUVs belong to continuous value spaces. In fact, a heuristic approach to apply DQN is to simply discretize the action spaces of AUVs as in \cite{sun2015target}, but a bad discretization cannot meet the requirement of control accuracy while a finer discretization will cause the curse of dimensionality. Based on DQN, deep deterministic policy gradient (DDPG), applying an actor network to approximate a deterministic policy \cite{silver2014deterministic}, was developed in \cite{lillicrap2015continuous} for continuous control, but DDPG only makes a compromise between the stability of learning and performance when applied to trajectory tracking problem of AUVs.

 Aiming at the aforementioned limitations of RL-based methods when considering underactuated AUVs with unknown dynamics and constrained inputs, this paper presents a model-free Multi Pseudo Q-learning based deterministic policy gradient (MPQ-DPG) algorithm to achieve high-level tracking control accuracy and stable learning based on a hybrid actors-critics architecture, where multiple actors and critics are trained to learn the deterministic policy and action-value function, respectively.

 Specifically, for the critics, the expected absolute Bellman error (EABE) is used to evaluate the performance of all critics in each time step, and the worst critic with maximum EABE is updated with a minibatch of transitions sampled randomly from a replay buffer \cite{lillicrap2015continuous}. This EABE-based updating rule can weaken the effect of bad critics and hence accelerate the convergence of learning. Moreover, due to the limitation of Q-learning that it is impractical to globally solve the greedy policy with optimization approaches in continuous action spaces, Pseudo Q-learning, which adopts the sub-greedy policy to replace the greedy policy in Q-learning, is developed to calculate the loss function with more accurate target value when updating the chosen critic. And in this sense, Pseudo Q-learning can be seen as an extension to Q-learning for problems with continuous action spaces. Based on Pseudo Q-learning, Multi Pseudo Q-learning (MPQ) is proposed to reduce the overestimation \cite{sutton2011reinforcement} resulting from the maximization in sub-greedy policy and to stabilize the learning without using target Q network as in DQN or DDPG. And relative to the target Q network, MPQ has two appealing advantages. Firstly, MPQ estimates the target value of critic more accurately and reduces the overestimation of action-value function. Secondly, the average operation makes the target value of critic more robust. For the actors, we randomly choose an actor to update in each time step, and similar to DDPG, the chosen actor is updated by applying the deterministic policy gradient. The final learned policy is defined as the average of all actors to avoid large but bad updates.

 The main contributions of this paper can be summarized as follows.
 \begin{itemize}
  \item To the best of our knowledge, this is the first time hybrid actors-critics architecture has been proposed and implemented to trajectory tracking problem of underactuated AUVs with unknown dynamics and constrained inputs. And multiple actors and critics are trained to learn the deterministic policy and action-value function.
  \item We propose Pseudo Q-learning, which adopts sub-greedy policy to replace greedy policy in Q-learning, to calculate the target value of critic. And Pseudo Q-learning can be seen as an extension to Q-learning for problems with continuous action spaces. The theoretical analysis demonstrates that Pseudo Q-learning makes it possible to calculate the loss function with more accurate target value when updating the critics.
  \item Inspired by Double Q-learning \cite{van2016deep}, Multi Pseudo Q-learning (MPQ) is proposed to reduce the overestimation resulting from the maximization in sub-greedy policy and to stabilize the learning. And the simulation results demonstrate that MPQ is a more effective way to stabilize the learning, compared to target Q network used in DQN and DDPG.
  \item The effectiveness and generality of MPQ-DPG is verified by the application on AUV with two different reference trajectories. And compared to DDPG and PIDNN control, MPQ-DPG shows great advantages with higher-level tracking control accuracy and better stability of learning.
 \end{itemize}

 The remainder of this paper is organised as follows. Section \ref{sec:formulation} describes the nonlinear model of underactuated AUVs with constraint inputs and trajectory tracking problem of AUVs. Section \ref{sec:MDP} introduces a new formulation for tracking control of AUVs under the framing of RL. In Section \ref{sec:algorithm}, the details of our proposed MPQ-DPG algorithm will be presented. In Section \ref{sec:simulation}, we demonstrate the simulation results by comparing with DDPG and PIDNN control on two different reference trajectories. Section \ref{sec:conclusion} draws the conclusions and outlines some directions for future work.

\section{Problem Formulation}\label{sec:formulation}
 In this section, we first give a brief description of classical 3-DOFs underactuated AUV models with input saturation. Then, trajectory tracking problem of AUVs is presented.
\subsection{Dynamics of AUV}
\begin{figure}[!t]
  \centering
  \scriptsize
    \includegraphics*[width=1.0\linewidth,viewport=210 232 640 372]{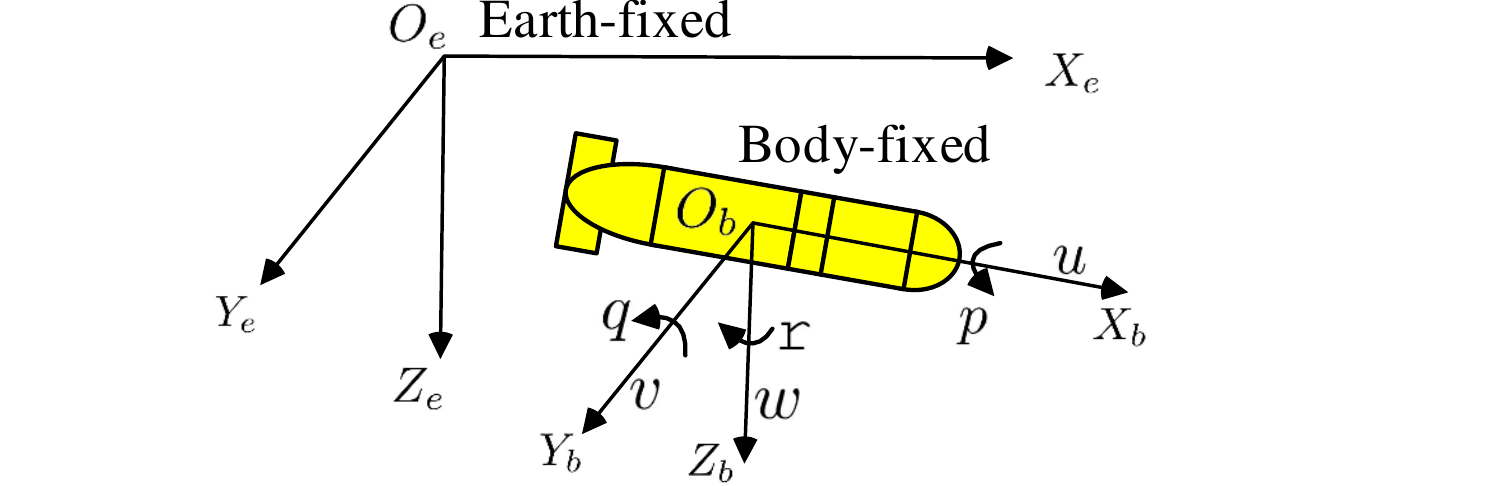}
  \caption{The earth-fixed and body-fixed reference frames for an AUV. The origin $O_b$ of the body-fixed frame is chosen to coincide with the center of gravity of the AUV.}
  \label{fig:AUV}
\end{figure}

 The study of dynamics of AUV can be divided into two parts, which comprise the motion and the forces causing this motion \cite{fossen2011handbook}. In general, the motion of AUVs involves 6-DOFs corresponding to the set of independent displacements which determine the position and orientation of AUV, as depicted in Fig. \ref{fig:AUV}, where $(u,v,w)$ are the surge, sway and heave velocities, $(p,q,\texttt{r})$ are the roll, pitch and yaw angular velocity. In this study, we consider the trajectory tracking problem only in the horizontal plane (lateral dynamics), and the model for the lateral dynamics of AUVs can be developed with a body-fixed coordinate frame $(u,v,\texttt{r})$ and an earth-fixed reference frame $(x,y,\psi)$, where $x$ and $y$ represent the coordinates of the AUV's center of mass, $\psi \in[-\pi,\pi]$ is the yaw angle.

 For simplicity, we make two common assumptions\cite{kim2015tracking}.
 \begin{enumerate}[(i)]
  \item The center of buoyancy coincides with the center of gravity of the AUV.
  \item The heave, pitch, and roll motions can be neglected.
 \end{enumerate}
 Then, the dynamics of AUV \cite{fossen2011handbook} are given as follows.
 \begin{align}
 \label{eqn:dynamics1}\dot{\et} & = \jb(\psi)\ph.  \\
 \label{eqn:dynamics2}\mb\dot{\ph}+\cb(\ph)\phi+\db(\ph)\ph & = \gb(\ph)\ta.
 \end{align}
 where $\et=[x,y,\psi]^{T}\in \mathbb{R}^3$ and $\ph=[u,v,\texttt{r}]^{T}\in \mathbb{R}^3$. $\mb\in \mathbb{R}^{3\times 3}$ and  $\cb(\ph) \in \mathbb{R}^{3\times 3}$ are the system inertia including added mass and the Coriolis-centripetal matrices, respectively. $\db(\ph) \in \mathbb{R}^{3\times 3}$ is the damping matrix. $\gb(\ph)\in \mathbb{R}^{3\times 2}$ is the input matrix. And $\jb(\psi)\in \mathbb{R}^{3\times 3}$ is the transformation matrix given as
 \begin{gather}
  \jb(\psi)=\begin{bmatrix}
      \cos\psi & -\sin\psi & 0  \\
      \sin\psi &  \cos\psi & 0  \\
      0        & 0        & 1
    \end{bmatrix}. \quad
 \end{gather}

 Clearly, the system models (\ref{eqn:dynamics1}) and (\ref{eqn:dynamics2}) are considered to be underactuated since $\ta=[\rm{sat}_{\bar{\xi}}(\xi),\rm{sat}_{\bar{\delta}}(\delta)]^{T}$ has only two independent input signals representing the saturated propeller thrust and rudder angle, respectively. $\rm{sat}_{\bar{\mu}}(\mu)$ stands for a saturation function defined as follows.
 \begin{equation}
  \rm{sat}_{\bar{\mu}}(\mu)=\left\{
   \begin{aligned}
    &\bar{\mu},  &&\rm{if}~\mu > \bar{\mu}  \\
    &\mu,        &&\rm{if}~|\mu| \leq \bar{\mu}  \\
    &-\bar{\mu}, &&\rm{if}~\mu < -\bar{\mu}  \\
   \end{aligned}.
  \right.
 \end{equation}
 where $\bar{\mu}$ denotes the saturation boundary for $\mu$.

 The model matrices $\mb,~\cb(\ph)$, $\db(\ph)$ and $\gb(\ph)$ are functions of the time-varying hydrodynamic coefficients \cite{fossen2011handbook}, which are related to the shape of AUV and hydrodynamic environment.

\subsection{Trajectory Tracking Problem of AUVs}
 Trajectory tracking problem of AUVs is to make the observed coordinate position $[x(t),y(t)]^T$ track a reference (desired) trajectory $\mathbf{d}(t)\in \mathbb{R}^2$ in an optimal manner. And the objective of trajectory tracking control of AUVs is to design an optimal tracking controller $\mathbf{\tau}^\ast$, which minimizes a predefined performance function $P(t_0,\tau)$ at initial time $t_0$. That is
 \begin{align}\label{eqn:objective}
  \mathbf{\tau}^\ast = \argmin_{\mathbf{\tau}} P(t_0,\tau).
 \end{align}

 For a reference trajectory $\mathbf{d}(t)=[x^d(t),y^d(t)]^T$, we define the tracking error as
 \begin{align}\label{eqn:tracking error}
  \mathbf{e}(t) = [x(t)-x^d(t),y(t)-y^d(t)]^T.
 \end{align}
 Then a general performance function $P(t_0,\tau)$ leading to the optimal tracking controller is defined as the following function of the tracking errors and control signals.
 \begin{align}\label{eqn:performance_C}
  P(t_0,\tau) = \int_{t_0}^{\infty}\gamma^{t-t_0}\big[\eb(t)^T\eb(t) +\mathbf{\tau}(t)^T\mathbf{H}\mathbf{\tau}(t)\big] dt.
 \end{align}
 where $\gamma$ is the discounting factor belonging to $(0,1]$, $\mathbf{H}\in \mathbb{R}^{2\times 2}$ is a symmetric positive definite matrix.

 Up to now, some traditional methods have been developed for the tracking control of underactuated AUVs with constrained inputs, but these methods either depend on accurate identification of the model or can not realize high-level control accuracy due to linearization or decoupling. In this paper, we present a new formulation of trajectory tracking problem of AUVs under the framing of RL to avoid these problems.

\section{A New Formulation under the Framing of Reinforcement Learning}\label{sec:MDP}
 The RL problem is meant to be a straightforward framing of the problem of learning from the interaction with an environment to achieve a goal. The major components of RL consist of agent, environment, state, action and reward. The learner or decision-maker is usually called the agent. The environment, comprising everything outside the agent, defines one instance of the reinforcement learning problem. The agent's sole objective is to find an optimal sequence of future actions (or control inputs) to maximize its cumulative rewards (or to minimize its cumulative control errors) from the current time step. The reward thus defines what are the good and bad actions for the agent.
 \begin{figure}[!b]
  \centering
  \scriptsize
    \includegraphics*[width=0.9\linewidth,viewport=160 195 690 400]{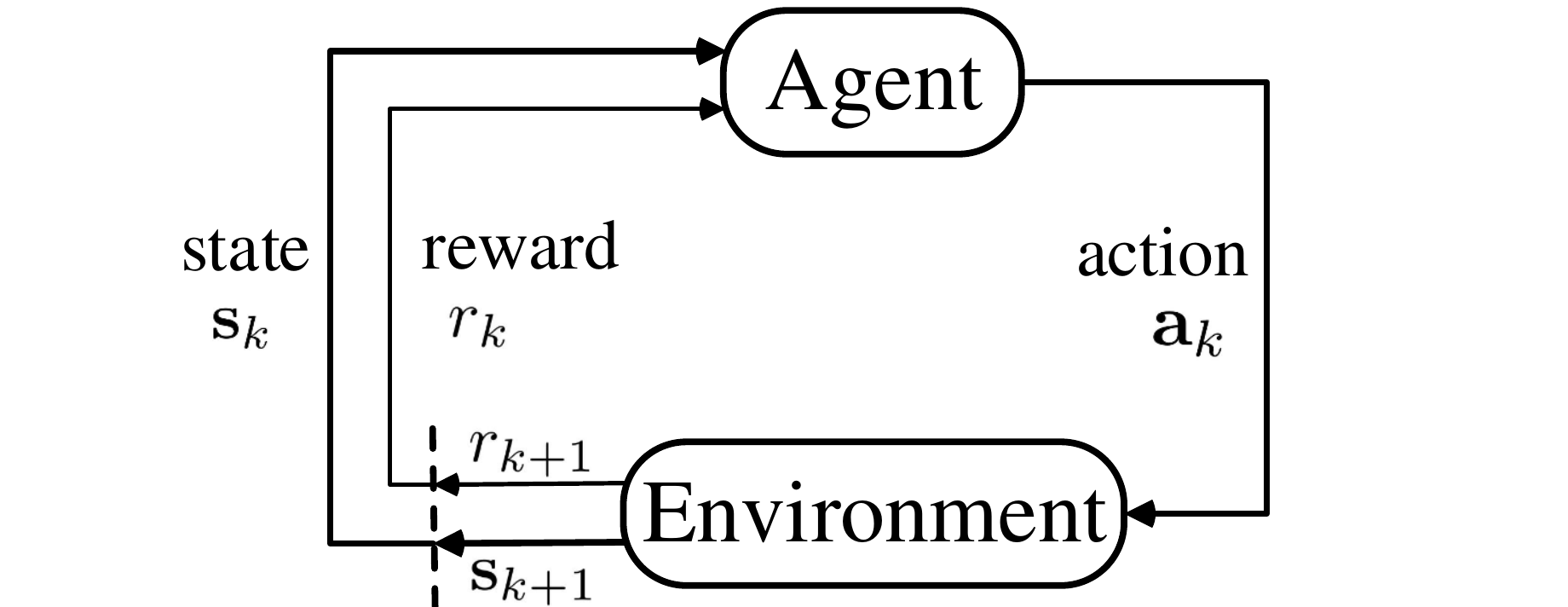}
  \caption{The agent-environment interaction in reinforcement learning.}
  \label{fig:agent_environment}
 \end{figure}

 Specifically, the agent and environment interact at each of a sequence of time steps. At $k$-th time step, the agent takes an action $\ab_k$ in state $\s_k$ and receives a scalar reward $r_{k+1}$, then the agent finds itself in a new state $\s_{k+1}$. Such a tuple ($\s_k,\ab_k,r_{k+1},\s_{k+1}$) is called a transition.  Fig. \ref{fig:agent_environment} diagrams the agent-environment interaction in RL \cite{sutton2011reinforcement}.

 To convert trajectory tracking problem of AUVs to a general RL problem, the continuous-time models (\ref{eqn:dynamics1}) and (\ref{eqn:dynamics2}) are discretized by the first-order Taylor expansion with sampling time $T_s$ as follows.
 \begin{align}
 \label{eqn:D-dynamics1} \et_{k+1} & = \et_k + T_s\jb(\psi_k)\ph_k.  \\
 \label{eqn:D-dynamics2} \ph_{k+1} & = \ph_k + T_s\mb^{-1}\mathbf{F}(\ta_k,\ph_k).
 \end{align}
 where $\mathbf{F}(\ta_k,\ph_k)=\gb(\ph_k)\ta_k-\cb(\ph_k)\ph_k-\db(\ph_k)\ph_k$. $\et_k,~\ph_k$, and $\ta_k$ are the sampled value of $\et,~\ph$, and $\ta$ at time $k\cdot T_s$. Meanwhile, corresponding to (\ref{eqn:performance_C}), we have the following discrete-time objective function $P_k$.
 \begin{align}\label{eqn:performance_D}
  P_k = \sum_{i\geq k}\gamma^{i-k}\big(\eb_i^T\eb_i+\mathbf{\tau}^T_i\mathbf{H}\mathbf{\tau}_i\big).
 \end{align}

 $Finite~Markov~Decision~Process$ ($finite$ MDP) is particularly important to the theory of RL. And the discrete-time models of AUV (\ref{eqn:D-dynamics1}) and (\ref{eqn:D-dynamics2}) can naturally be modeled as a $finite$ MDP dy designing the state and action as
 \begin{align}
 \label{eqn:states}  & \s_k = \big[\et^T_k,\ph^T_k,\mathbf{d}^T_k,\mathbf{d}_{k+1}^T\big]^T.  \\
 \label{eqn:actions} & \ab_k = \ta_k.
 \end{align}
 It should be mentioned that, the designed state $\s_k$ is based on the assumption that all elements in $\s_k$
 are measurable.

 In RL, the return is defined as the sum of discounted future rewards $R_k=\sum_{i=k}^{T}\gamma^{(i-k)}r(\s_i,\ab_i)$ with a discounting factor $\gamma \in (0,1]$. An agent's behavior is defined by a stochastic policy $\pi:\mathcal{S}\rightarrow \mathcal{P}(\mathcal{A})$. The action-value function describes the expected return after taking an action $\ab_k$ following policy $\pi$ in state $\s_k$ and thereafter is defined as:
 \begin{align}\label{eqn:action-value function}
  Q^{\pi}(\s_k,\ab_k) = \mathbb{E}_{r_{i > k}, \s_{i > k} \sim E, \ab_{i > k} \sim \pi}[R_k|\s_k,\ab_k].
 \end{align}
 Then the agent's goal is to learn a policy which maximizes the expected return from the start distribution, denoted by $J$.
 \begin{align}\label{eqn:expected return}
  J = \mathbb{E}_{r_i, \s_i \sim E, \ab_i \sim \pi}[R_0]=\mathbb{E}_{p(\s_0), \ab_0 \sim\pi}\left[Q^{\pi}(\s_0,\ab_0)\right].
 \end{align}
 Hence, the core of RL is how to solve the policy $\pi$ and action-value function $Q^{\pi}(\s_0,\ab_0)$.

 How to design the reward function is one of the most important problem in RL, whether the reward function is good empirically determines the performance of controller. Actually, the idea behind RL-based method for trajectory tracking problem of AUVs is to make the agent's goal (\ref{eqn:expected return}) match the objective (\ref{eqn:objective}) of trajectory tracking control. To this end, the reward function should be designed to make the expected return $R_k$ equivalent to the performance function $P_k$ (\ref{eqn:performance_D}). And based on these considerations, the reward function $r_{k+1}$ is designed as
 \begin{align}\label{eqn:reward}
  r_{k+1} = r(\s_k, \ab_k) = -\left(\eb_k^T\eb_k+\ab^T_k\mathbf{H}\ab_k\right).
 \end{align}

 Note that model matrices $\mb,~\cb(\ph)$, $\db(\ph)$ and $\gb(\ph)$ are difficult to be identified in real application. Fortunately, these matrices are not required when designing basic components such as state $\s_k$, action $\ab_k$ and reward $r_{k+1}$, which are the only things need to be known about AUVs in RL.

 However, many existing RL algorithms are only tested on some relatively simple control tasks such as cart-pole, swing-up and mountain-car, which have not very high requirements of the control accuracy and the stability of learning. But these two performance indexes are significant for the trajectory tracking problem of AUVs. For these purposes, we develop a novel RL algorithm to guarantee high-level control accuracy and stable learning in next section.

\section{Multi Pseudo Q-learning based Deterministic Policy Gradient Algorithm} \label{sec:algorithm}
 In this section, a novel architecture with multiple actors and critics will be first presented to approximate the deterministic policy and action-value function. Then, the strategies for updating the critics and actors are developed. Next, we propose an innovative approach called Multi Pseudo Q-learning (MPQ) to calculate the target value of critics. Finally, the stability analysis of the learning is given.
\subsection{Hybrid Actors-Critics Architecture}
 We consider a standard reinforcement learning problem in which an agent interacts with a stochastic environment $E$ by sequentially choosing actions in discrete time steps. And we model it as a $finite$ MDP which comprises: a state space $\mathcal{S}$, an action space $\mathcal{A}$, an initial state distribution $p(\s_0)$, a stationary transition dynamics distribution $p(\s_{k+1}|\s_k,\ab_k)$ satisfying the Markov property, and a reward function $r_{k+1}=r(\s_k,\ab_k)$.

 Similar to dynamic programming, many RL methods make use of the following recursive Bellman equation \cite{sutton2011reinforcement} to recursively solve the action-value function.
 \begin{equation}\label{eqn:Q-function}
  \begin{split}
   Q^{\pi}\!(\s_k, \ab_k) \!=\! \mathbb{E}_{r_{k+1}\!, \s_{k+1} \sim E}&\!\big[r_{k+1} +  \\
   &\! \gamma \mathbb{E}_{\ab_{k+1}\sim \pi}[Q^{\pi}\!(\s_{k+1}, \ab_{k+1})]\big].
  \end{split}
 \end{equation}

 And if the target policy is deterministic, we can denote it as a function $\mu:\mathcal{S}\rightarrow \mathcal{A}$, then avoid the inner expectation:
 \begin{equation}\label{eqn:Determinnistic Bellman equation}
  \begin{split}
   Q^\mu(\s_k,\ab_k) = \mathbb{E}_{r_{k+1}, \s_{k+1} \sim E}\big[&r_{k+1} +  \\
   & \gamma Q^\mu\big(\s_{k+1},\mu(\s_{k+1})\big)\big].
  \end{split}
 \end{equation}
 Meanwhile, for a deterministic policy $\mu$, the expectation about $\ab_0$ in (\ref{eqn:expected return}) is also disappeared.
 \begin{align}\label{eqn:D-expected return}
  J = \mathbb{E}_{p(\s_0)}\big[Q^\mu(\s_0,\mu(\s_0))\big].
 \end{align}

\begin{figure}[!t]
  \centering
  \subfigure[Actor network]{
   \scriptsize
    \includegraphics*[width=0.9\linewidth,viewport=210 255 620 418]{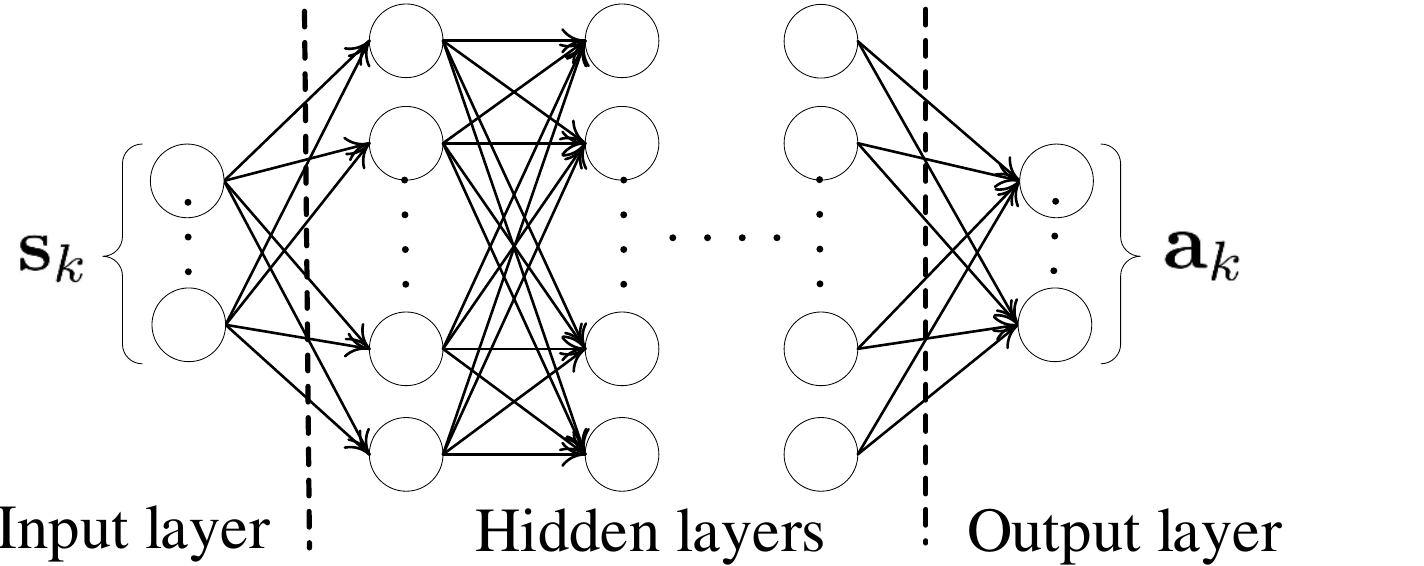}}
    \label{fig:subfig:a}
  \hspace{1in}
  \subfigure[Critic network]{
   \scriptsize
    \includegraphics*[width=0.9\linewidth,viewport=235 195 650 400]{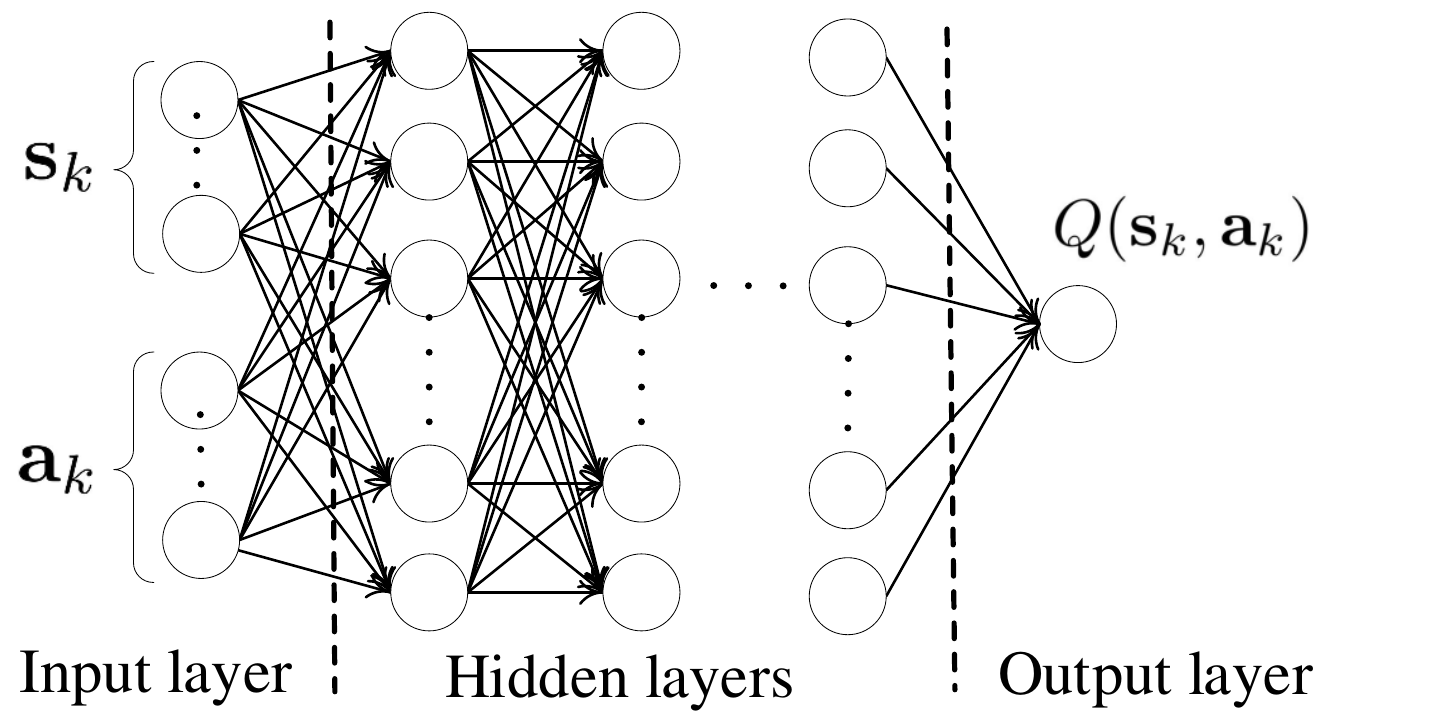}}
    \label{fig:subfig:b}
  \caption{Illustration of the structures of actor network and critic network.}
  \label{fig:NN}
 \end{figure}

 Different from many existing policy gradient methods adopting the general single actor-critic architecture to learn approximations to both policy and action-value function according to (\ref{eqn:Determinnistic Bellman equation}) and (\ref{eqn:D-expected return}), a novel hybrid actors-critics architecture with $n$-actors and $m$-critics is employed to realize high-level tracking control accuracy of AUVs and stable learning. Specifically, We construct $n$ actors $\mu_{\mathbf{\theta}_i} (\s)$ ($i=1,\cdots,n$) to learn the optimal deterministic policy and $m$ critics $Q_{\mathbf{\omega}_j}(\s,\ab)$ ($j=1,\cdots,m$) to learn the action-value function which evaluates the current policy learned by the actors. $\mathbf{\theta}_i$ and $\mathbf{\omega}_j$ are the parameters of the $i$-th actor and the $j$-th critic, respectively. Due to the complexity of underactuated AUVs model, fully connected DNN is adopted to improve the approximation capability of actors-critics. The structures of actor network and critic network are depicted in Fig. \ref{fig:NN}.

 Under the framework of hybrid actors-critics, the final learned policy $\mu_f(\s_k)$ is the average of $n$ actors as follows:
 \begin{align}\label{eqn:average policy}
  \ab_k=\mu_f(\s_k)=\frac{1}{n}\sum_{i=1}^{n}\mu_{\mathbf{\theta}_i}(\s_k).
 \end{align}
 Note that this design is not the major factor influencing the learning speed due to two facts. First, only one actor is learned in each time step. Second, the optimality of actors depends on the optimality of critics since the updating of actors is based on estimation of action-value function by critics, hence the number of critics is main factor influencing the learning speed.

\begin{figure}[!t]
  \centering
   \scriptsize
    \includegraphics*[width=0.9\linewidth,viewport=210 240 640 390]{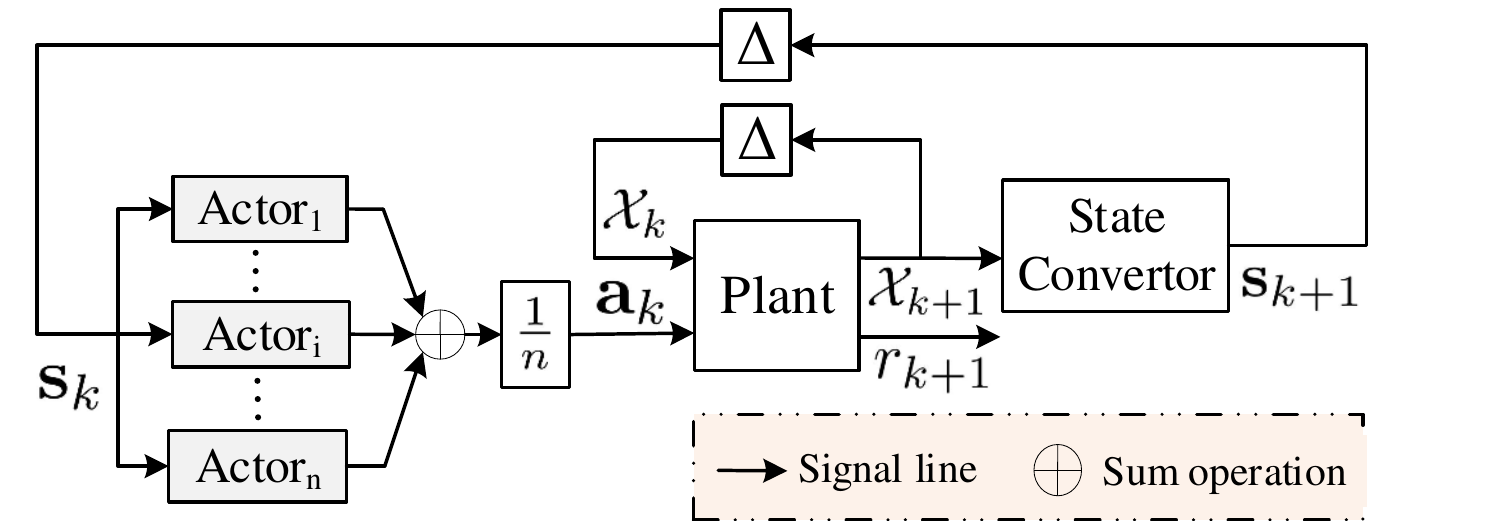}
  \caption{Block diagram of the average policy $\mu_f(\s_k)$ for an AUV system. $\text{Actor}_{i}$ denotes the $i$-th actor. $\mathcal{X}_{k}$ denotes the observation states of the system at time $k\cdot T_s$. And $\Delta$ is the time delay module.}
  \label{fig:dynamic}
 \end{figure}

 In earlier works such as DDPG, there is only one actor, thus the policy learned by the actor will change dramatically once some bad training samples are encountered during the training. By contrast, the average policy (\ref{eqn:average policy}) can avoid this problem. And Fig. \ref{fig:dynamic} shows the block diagram of our proposed average policy for the AUV system.

\subsection{Update Strategies for Actors-Critics Network}
 Note that the expectation in (\ref{eqn:Determinnistic Bellman equation}) depends only on the environment $E$, which makes it possible to learn actors and critics off-policy using transitions sampled from a replay buffer, which is a finite sized cache $\mathcal{R}$ filled with previous transitions. And the transitions in $\mathcal{R}$ can be seen as generated from a different stochastic behavior policy $\beta$, which will be used henceforth in this paper.

 For multiple critic networks, a simple updating rule is to update the worst one with maximum expected absolute Bellman error (EABE) among all critics in each time step. And the EABE of the $j$-th critic is defined as follows.
 \begin{equation}\label{eqn:EABE}
   \begin{split}
    \text{EABE}_j \!=\! \mathbb{E}_{r_{k+1}\sim E, \s_k \sim \rho^\beta,\ab_k\sim\beta}&\big| Q_{\mathbf{\omega}_j}(\s_k,\ab_k)-r_{k+1}-  \\
    & \gamma Q_{\mathbf{\omega}_j}(\s_{k+1},\mu_\mathbf{\theta}(\s_{k+1}))\big|.
   \end{split}
 \end{equation}
 where $\mu_\mathbf{\theta}(\s)$ is the actor updated in the last time step. Then, the selected worst critic $c = \argmax_j\text{EABE}_j$ is optimized by minimizing the following loss function:
 \begin{align}\label{eqn:loss}
  L(\mathbf{\omega}_c) \!=\! \mathbb{E}_{r_{k+1}\sim E, \s_k \sim \rho^\beta,\ab_k\sim\beta} \bigg[\big(Q_{\mathbf{\omega}_c}(\s_k,\ab_k) \!-\! Y^c_{k+1}\big)^2\bigg].
 \end{align}
 where $Y^c_{k+1}$ is the target value of $Q_{\mathbf{\omega}_c}(\s_k,\ab_k)$ and has the following form.
 \begin{align}\label{eqn:target value}
  Y^c_{k+1} &= r_{k+1}+\gamma Q(\s_{k+1},\ab_{k+1}).
 \end{align}

 As in DQN, one way to estimate $Y^c_{k+1}$ is to combine the target Q network with Q-learning, which is an off-policy temporal-difference algorithm using the greedy policy $\mu_g(\s)$.
 \begin{align}
  \label{eqn:greedy policy} \mu_g(\s_{k+1})&=\argmax_\ab Q_{\mathbf{\omega}^\prime_c}(\s_{k+1},\ab). \\
  \label{eqn:target Q network} Q(\s_{k+1},\ab_{k+1}) &= Q_{\mathbf{\omega}^\prime_c}(\s_{k+1},\mu_g(\s_{k+1})).
 \end{align}
 where $Q_{\mathbf{\omega}^\prime_c}$ is the target Q network of the $c$-th critic network $Q_{\mathbf{\omega}_c}$, and $\mathbf{\omega}^\prime_c$ is updated by slowly tracking $\mathbf{\omega}_c$.

 However, it is impossible to directly apply Q-learning in continuous domains, because the optimization of action $\ab$ in (\ref{eqn:greedy policy}) at every time step is too slow to be practical with fully connected DNN approximators and nontrivial, continuous action spaces of AUVs. Instead, a novel approach called Multi Pseudo Q-learning (MPQ) will be proposed in the next subsection to estimate target value $Y^c_{k+1}$ more accurately and to avoid the global maximization in (\ref{eqn:greedy policy}).

 Different from the critics, the actors are updated by using the deterministic policy gradient defined as the gradient of the policy's performance (\ref{eqn:D-expected return}). And the deterministic policy gradient is usually estimated by applying the chain rule to the expected return from the start distribution $J$ with respect to the actor parameters $\mathbf{\theta}_i$ as follows \cite{silver2014deterministic}.
 \begin{equation}\label{eqn:policy gradient}
  \begin{split}
   \nabla_{\mathbf{\theta}_i}J &\approx \mathbb{E}_{\s_k\sim \rho^\beta}\big[\nabla_{\mathbf{\theta}_i}Q_{\mathbf{\omega}_c}(\s_k,\ab)|_{\ab=\mu_{\mathbf{\theta}_i}(\s_k)}\big]\\
   &= \mathbb{E}_{\s_k\sim \rho^\beta}\big[\nabla_\ab Q_{\mathbf{\omega}_c}(\s_k,\ab)|_{\ab=\mu_{\mathbf{\theta}_i}(\s_k)}\nabla_{\mathbf{\theta}_i}\mu_{\mathbf{\theta}_i}(\s_k)\big].
  \end{split}
 \end{equation}

 Note that the actor to be learned is selected randomly among all actors in each time step, that is, each actor will be updated with equal probability $1/n$.

 \begin{figure*}[!t]
  \centering
  \scriptsize
    \includegraphics*[width=0.8\linewidth,viewport=192 175 657 423]{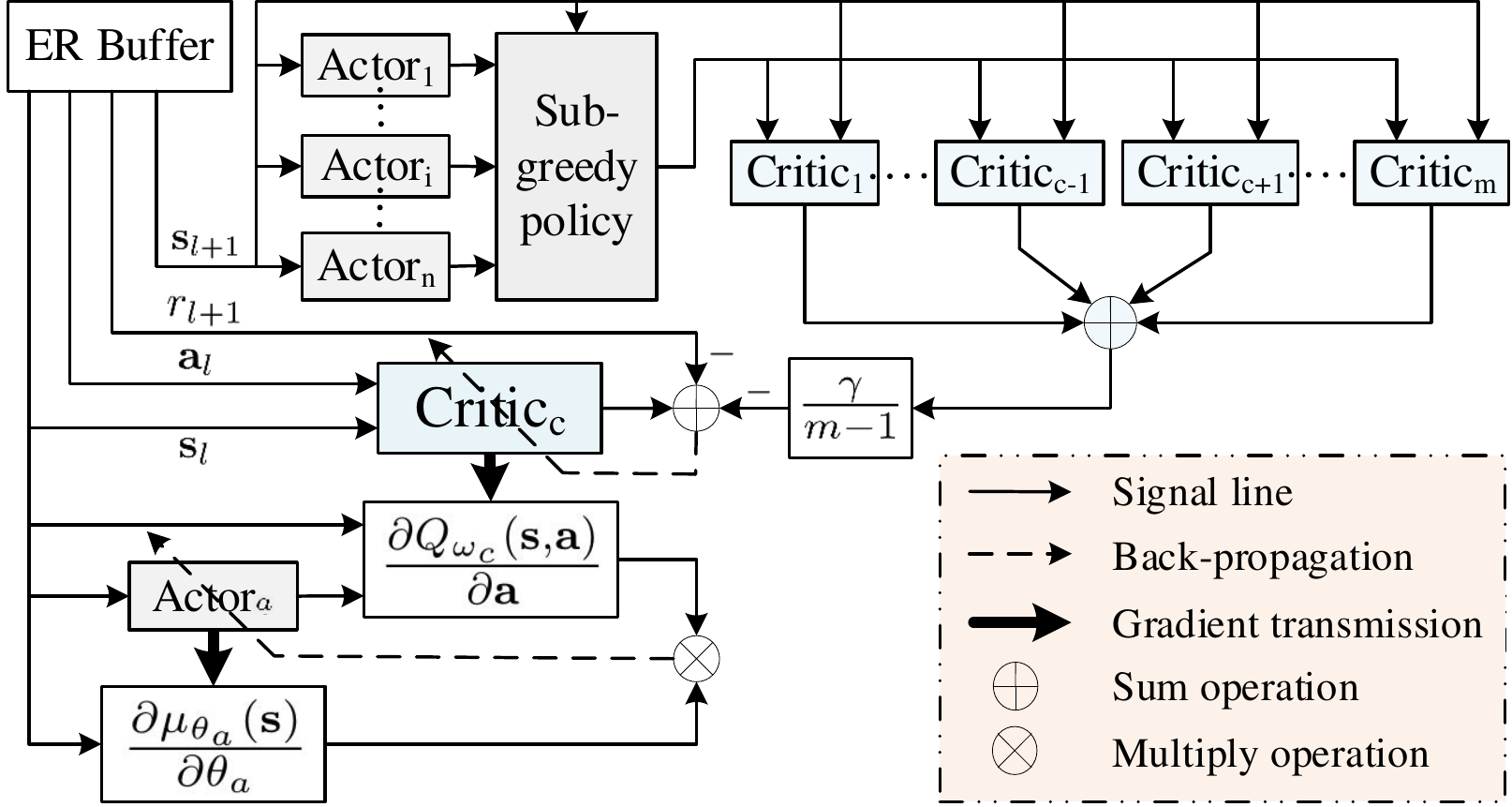}
  \caption{Block diagram of the MPQ-DPG algorithm. ER Buffer is the experience replay buffer. $\text{Actor}_{i}$ and $\text{Critic}_{j}$ denote the $i$-th actor and the $j$-th critic, respectively. $c$ is chosen according to the EABE-based updating rule of critics, and $a$ is chosen randomly in each iteration step.}
  \label{fig:update}
 \end{figure*}

\subsection{Multi Pseudo Q-learning}
 Many existing policy optimization methods such as DDPG have the following two main drawbacks, and which will be exacerbated when applied to the trajectory tracking control of underactuated AUVs due to strong coupling and severe nonlinearity of AUVs model.
 \begin{enumerate}[(i)]
  \item The critic is usually inefficient to evaluate the policy learned by the actor.
  \item The action-value function learned by critic has high variance.
 \end{enumerate}

 A reasonable way to view these problems is that there exists large bias between the action-value function $Q(\s,\ab)$ learned by current critic and $Q^{\mu}(\s,\ab)$ following the current learned policy $\mu$ in DDPG, and this bias will result in biased target value and thus make the loss function (\ref{eqn:loss}) incapable to measure the performance of the critic. Indeed this bias is inevitable when the approximator is used for the action-value function, but we can mitigate the effect of this bias by using the greedy policy to determine the next action $\ab_{k+1}$ in (\ref{eqn:target value}) to partially correct the biased target value. More essentially, from the perspective of the Bellman optimality equation, it is necessary to use the greedy policy for calculating the target value $Y^c_{k+1}$ due to its unique advantage that the greedy policy will evolve into the optimal policy once the optimal action-value function $Q^*(\s,\ab)$ is found by the critic according to the following Bellman optimality equation:
 \begin{equation}
  \begin{split}
   Q^*(\s_k,\ab_k) = \mathbb{E}_{r_{k+1},\s_{k+1}\sim E}\big[&r_{k+1} +  \\
   & \gamma\max_{\ab}~Q^*(\s_{k+1},\ab)\big].
  \end{split}
 \end{equation}

 In this sense, although it is impractical to solve globally the greedy policy with the optimization methods in continuous action spaces, a compromise scheme is still to estimate $Q(\s_{k+1},\ab_{k+1})$ with a policy close enough to the greedy policy while avoiding the global maximization. To this end, we calculate the target value $Y^c_{k+1}$ with a new approach called Pseudo Q-learning, which adopts the following sub-greedy policy $\mu_{sg}(\s|Q_{\mathbf{\omega}_j})$ to replace the greedy policy $\mu_g(\s)$ in (\ref{eqn:target Q network}).
 \begin{align}\label{eqn:sub-greedy policy}
  \mu_{sg}(\s|Q_{\mathbf{\omega}_j})=\argmax_{\ab\in \mathcal{A}(\s)} Q_{\mathbf{\omega}_j}(\s,\ab).
 \end{align}
 where $\mathcal{A}(\s) = \{\mu_{\mathbf{\theta}_i}(\s),i=1,\cdots,n\}$ is the set of actions obtained from all actors in state $\s$. Actually, Pseudo Q-learning can be seen as an extension of Q-learning to RL problems with continuous action spaces in the sense that the sub-greedy policy will gradually approach the greedy policy as the learning proceeds. And note that the sub-greedy policy avoids global optimization.

 However, similar to Q-learning, there also exists a maximization bias or overestimation in Pseudo Q-learning, and this is due to the fact that same samples are used both to determine the maximizing action and to estimate its value. To reduce this overestimation, we apply the idea underlying Double Q-learning \cite{van2016deep} to develop Multi Pseudo Q-learning (MPQ), which decouples the selection of the next action and the estimation of its value by using multiple critics. Specifically, at each time step, MPQ determines the next action $\ab_{k+1}$ conditioned on the selected critic $Q_{\mathbf{\omega}_c}$ according to the sub-greedy policy (\ref{eqn:sub-greedy policy}), and then calculates $Q(\s_{k+1},\ab_{k+1})$ as the average of all other critics $Q_{\mathbf{\omega}_j} (j\neq c)$. Consequently, a more accurate target value $Y^c_{k+1}$ is developed as follows.
 \begin{align}
  \label{eqn:next action} \ab_{k+1} &= \mu_{sg}(\s_{k+1}|Q_{\mathbf{\omega}_c}). \\
  \label{eqn:state value} Y^c_{k+1} &= r_{k+1}+\frac{\gamma}{m-1}\sum_{j=1,j\neq c}^{m}Q_{\mathbf{\omega}_j} \big(\s_{k+1},\ab_{k+1}\big).
 \end{align}

 \begin{algorithm}[!t]
    \caption{MPQ-DPG Algorithm}
    \label{alg:MPQ-DPG}
    Initialize $m$ critics $Q_{\mathbf{\omega}_j}(\s,\ab)$ ($j=1,\cdots,m$) and $n$ actors $\mu_{\mathbf{\theta}_i} (\s)$ ($i=1,\cdots,n$) with random weights $\mathbf{\omega}_j$ and $\mathbf{\theta}_i$\;
    Initialize replay buffer $\mathcal{R}$\ and $a = \text{randint}\big(1,[1,n]\big)$\;
    \For{episode = 1, M}
    {
      Initialize a random noise process $\mathcal{N}$ for exploration\;
      Receive initial observation state $\s_0$\;
      \For{$k = 0, K$}
      {
        Select action $\ab_k=\frac{1}{n}\sum_{i=1}^{n}\mu_{\mathbf{\theta}_i}(\s_k)+\mathcal{N}_k$ according to $n$ actors and exploration noise\;
        Execute action $\ab_k$ and receive $r_{k+1}$, $\s_{k+1}$\;
        Store transition ($\s_k,\ab_k,r_{k+1},\s_{k+1}$) in $\mathcal{R}$\;
        Sample a random minibatch of $N$ transitions ($\s_l,\ab_l,r_{l+1},\s_{l+1}$) from $\mathcal{R}$\;
        Calculate the sampled Bellman absolute error:
        \resizebox{.8\hsize}{!}{$~~~~~~\text{EABE}_j = \frac{1}{N}\sum\limits_{l}\bigg| Q_{\mathbf{\omega}_j}(\s_l,\ab_l)-r_{l+1}-$}\
        \resizebox{.9\hsize}{!}{$~~~~~~~~~~~~~~~~~~~~~~~~~\gamma Q_{\mathbf{\omega}_j}(\s_{l+1},\mu_{\mathbf{\theta}_a}(\s_{l+1}))\bigg| $}\;
        Select the worst critic $c = \argmax_j\text{EABE}_j$\;
        Determine the next action $\ab_{l+1}$ according to (\ref{eqn:next action})\;
        Calculate the target value $Y^c_{l+1}$ according to (\ref{eqn:state value})\;
        Update the $c$-th critic by minimizing the loss:
        $~~~~~~~L(\mathbf{\omega}_c)=\frac{1}{N}\sum\limits_{l}\big(Q_{\mathbf{\omega}_c}(\s_l,\ab_l)-Y^c_{l+1}\big)^2$\;
        Reset $a = \text{randint}\big(1,[1,n]\big)$\;
        Update the $a$-th actor using the sampled policy gradient:
        $~~~\nabla_{\mathbf{\theta}_a}J\!=\!\frac{1}{N}\!\sum\limits_{l}\nabla_\ab Q_{\mathbf{\omega}_c}(\s_l,\ab)|_{\ab=\mu_{\mathbf{\theta}_a}(\s_l)}\nabla_{\mathbf{\theta}_a}\mu_{\mathbf{\theta}_a}(\s_l)$\;
      }
    }
 \end{algorithm}

 What deserves special attention about MPQ is that the critic $Q_{\mathbf{\omega}_c}$ being updated is different from the critics being used to calculate the target value $Y^c_{k+1}$. Therefore, the stability of learning can be guaranteed without the help of the target Q network (\ref{eqn:target Q network}), which is used in DQN and DDPG to effectively alleviate the issue of unstable learning.

 Our model-free algorithm, which we call Multi Pseudo Q-learning based Deterministic Policy Gradient (MPQ-DPG, Algorithm \ref{alg:MPQ-DPG}), updates critics and actors based on MPQ and DPG, respectively. And the structure diagram of the MPQ-DPG Algorithm is shown in Fig. \ref{fig:update}.

\subsection{Stability Analysis of the Learning}
 Prior to DQN, it was often not suggested to utilize large, nonlinear function approximators for learning action-value or value functions due to the drawback that the learning is often unstable. However, such function approximators often appear essential to deal with many challenging problems where useful features can not be handcrafted, or where the state is partially observed and high-dimensional. For the trajectory tracking problem of underactuated AUVs considered in this paper, three innovations are employed to make it possible to train the full connected DNN function approximators in a stable way. First, the actors and critics are trained off-policy with transitions sampled from a replay buffer to minimize the correlations between transitions. Second, the final learned policy is defined as the average of all actors to avoid large but bad update of policy. Third, a more accurate and robust target value of critic is obtained with MPQ approach when updating the critics.

 As a contrast, for many complex and practical applications such as the trajectory tracking problem of underactuated AUVs, the target Q network used in DQN and DDPG can not satisfy the high requirement of the stability of learning. But our proposed MPQ approach can realize more stable learning due to the following three observations.
 \begin{enumerate}[(i)]
  \item When bad transitions are sampled from replay buffer $\mathcal{R}$, only one critic will be affected. And If there is a drastic change in one $Q$ function, its effect on other $Q$ functions will be averaged.
  \item The average operation in MPQ makes the estimation of the target value $Y^c_{k+1}$ more robust, thus stabilizes the learning and reduces the variance of the action-value function learned by the critics.
  \item Once the optimal action-value function is learned by the critics, the loss function of critics will be close to zero, and the update of critics will tend to be slow and stable.
 \end{enumerate}
 Due to the above advantages over the target Q network, MPQ approach is more effective to stabilize the learning when a substantial of noise or model uncertainties exist, especially in the underwater environment.

\section{Simulation studies} \label{sec:simulation}
 In this section, the proposed MPQ-DPG algorithm will be implemented to the trajectory tracking problem of a classical and widely used AUV called REMUS \cite{prestero2001verification}, which is top-bottom (xy-plane) and port-starboard (xz-plane) symmetry. And the model matrices $M$, $C(\phi)$, $D(\phi)$ and $G(\phi)$ of REMUS are given in Appendix \ref{app:model coefficients}.

 Consider the AUV models (\ref{eqn:dynamics1}) and (\ref{eqn:dynamics2}) in which the input saturation boundaries $\bar{\xi}=86~\text{N}$, $\bar{\delta}=13.6\times \pi/180~\text{rad}$ and the hydrodynamic coefficients are listed in Appendix \ref{app:model coefficients}. The weight matrix in reward function (\ref{eqn:reward}) is chosen as $\mathbf{H}=0.001I_{2\times2}$. Moreover, to demonstrate the generality of our proposed MPQ-DPG algorithm, two different reference trajectories of REMUS are adopted as follows:
\begin{enumerate}[ 1)]
  \item Reference trajectory 1 (RT1):
    \begin{align}
     &x^d = (15-0.1t)\cos(\frac{\pi}{20}t). \nonumber\\
     &y^d = (15-0.1t)\sin(\frac{\pi}{20}t). \nonumber
    \end{align}
  \item Reference trajectory 2 (RT2):
    \begin{align}
     &x^d = 0.8t-40. \nonumber\\
     &y^d = 10\sin(\frac{\pi}{25}t).\nonumber
    \end{align}
 \end{enumerate}
 Actually, these two reference trajectories have practical significance in various underwater applications. RT1 is an asymptotic helical curve widely used in target monitoring and approaching. And RT2 is sine curve widely used in obstacle avoidance.

 The reinforcement interval $T_s$ is chosen as $0.1 s$, and the reference trajectory is tracked in 100 seconds, which means there are 1000 time steps in each episode. Moreover, all simulations are run with 1500 episodes, and in each episode, the initial positions $(x_0,y_0)$ are randomly located in domain $\big([14,16], [-1,1]\big)$ and domain $\big([-41,-39], [-1,1]\big)$ for RT1 and RT2, respectively. The initial yaw angle is both randomly chosen from $[\pi/4,3\pi/4]$. And the initial velocities $(u_0,v_0,r_0)$ are both randomly chosen from $\big([1,1.5], [-0.3,0.3],[-0.2,0.2]\big)$.
 \begin{figure*}[!t]
  \centering
  \begin{minipage}[t]{0.5\linewidth}
   \centering
   \includegraphics[width=0.82\linewidth,viewport=50 240 550 600]{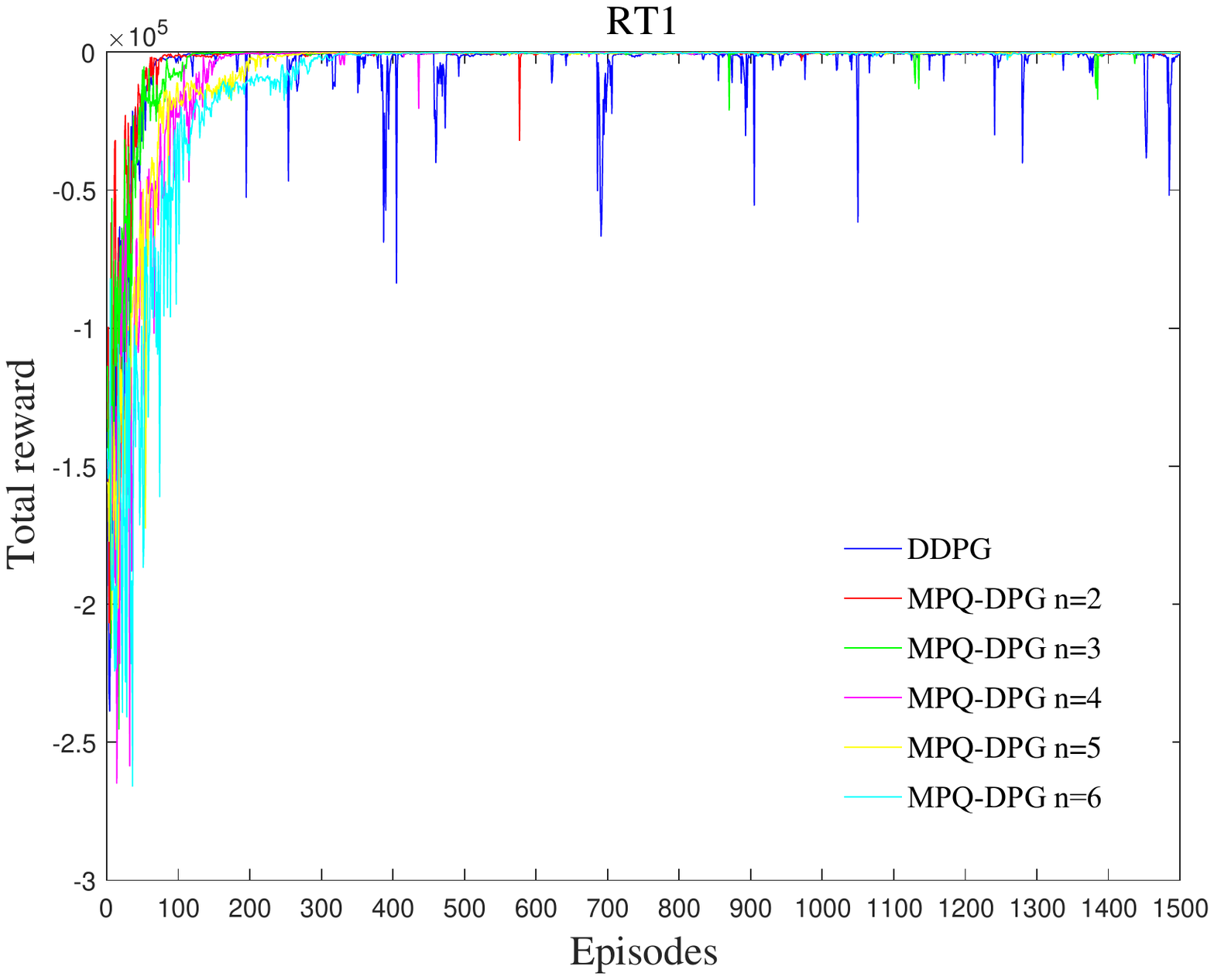}
  \end{minipage}%
  \begin{minipage}[t]{0.5\linewidth}
   \centering
   \includegraphics[width=0.82\linewidth,viewport=50 240 550 600]{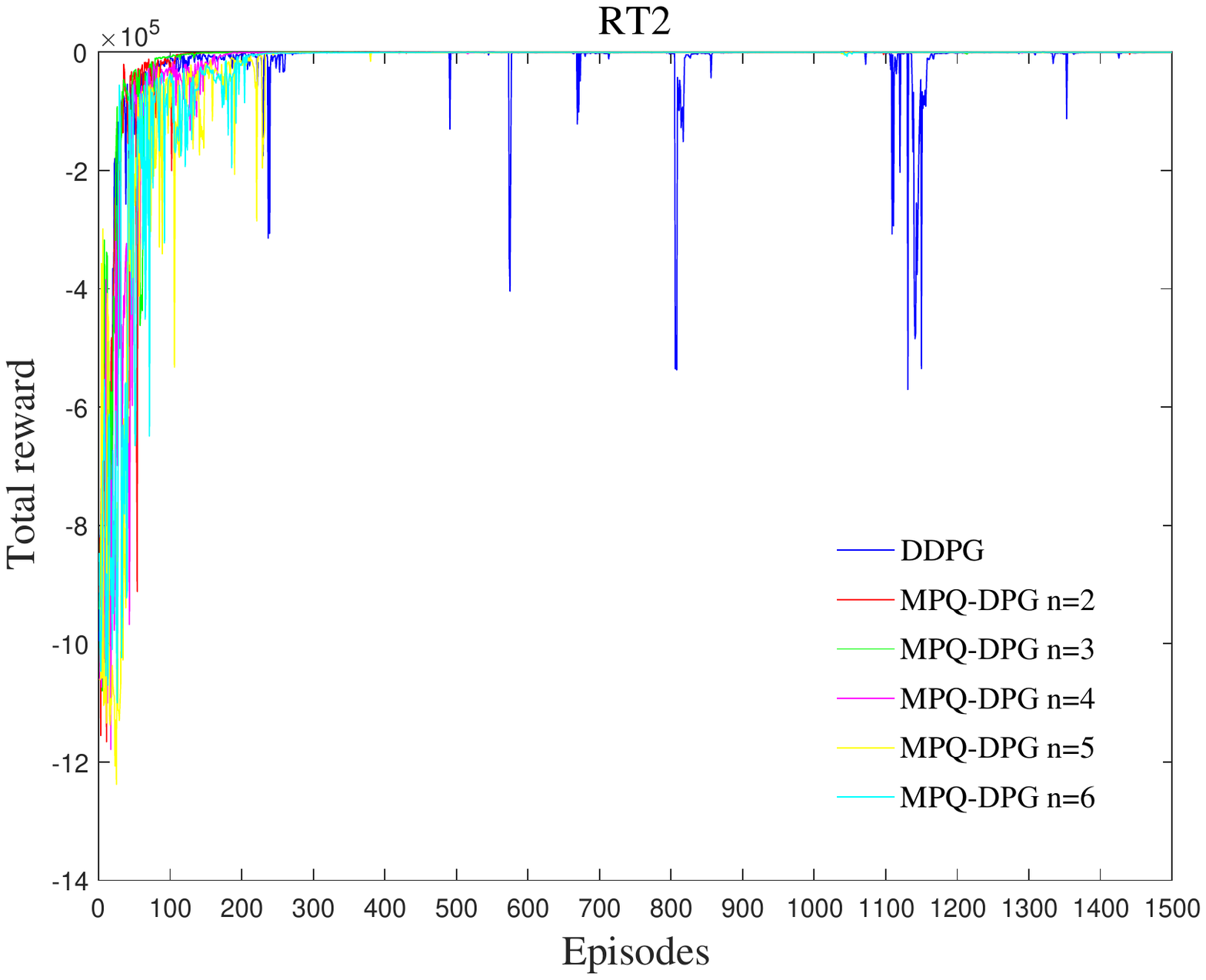}
  \end{minipage}
  \caption{DDPG and MPQ-DPG ($n=2,3,4,5,6$) learning curves in one learning trial with different reference trajectories. The left and right figures show the results corresponding to RT1 and RT2, respectively. The learning curves demonstrate that MPQ-DPG with $n\geq 2$ outperforms DDPG regarding the stability of learning process.}
  \label{fig:fig1}
 \end{figure*}
 \begin{figure*}[!ht]
  \centering
  \begin{minipage}[t]{0.5\linewidth}
   \centering
   \includegraphics*[width=0.82\linewidth,viewport=50 230 550 620]{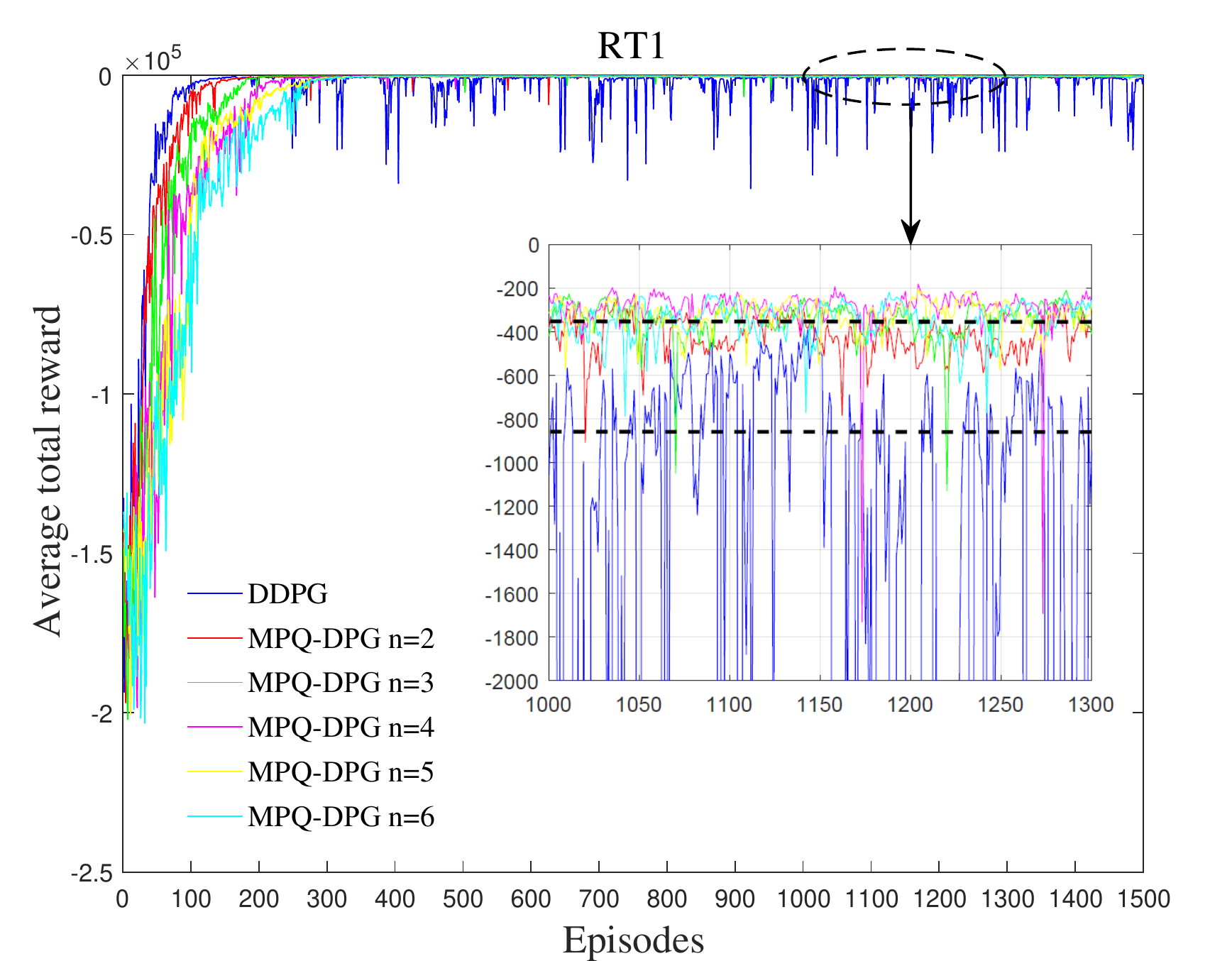}
  \end{minipage}%
  \begin{minipage}[t]{0.5\linewidth}
   \centering
   \includegraphics*[width=0.82\linewidth,viewport=50 230 550 620]{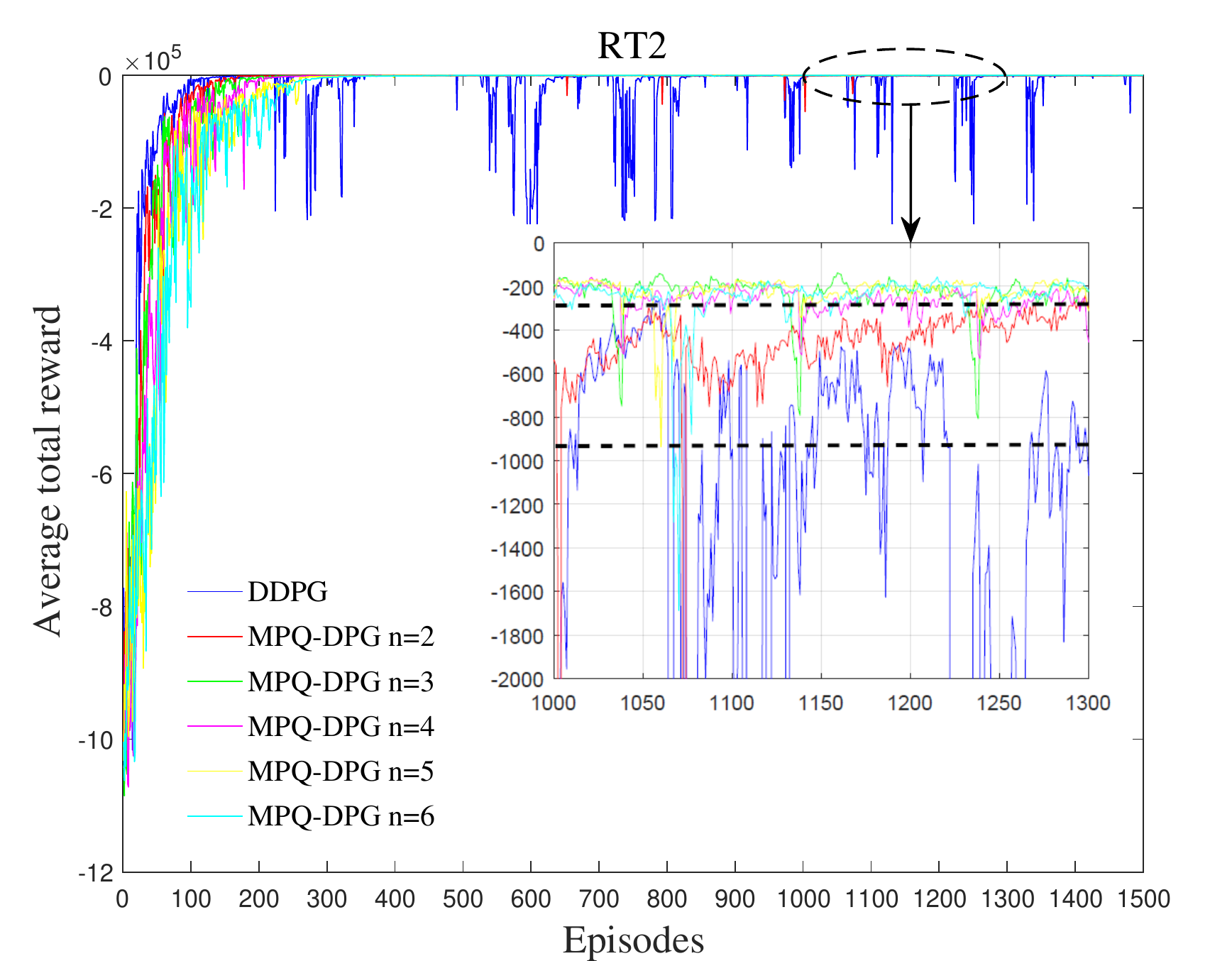}
  \end{minipage}
  \caption{DDPG and MPQ-DPG ($n=2,3,4,5,6$) performance (average total reward) curves of different reference trajectories. All curves are averages over five independent learning trials with different random seeds. The left and right figures show the results corresponding to RT1 and RT2, respectively. And the \textbf{black dashed lines} in local enlarged drawings of chosen stable intervals indicate the average levels, which demonstrate that MPQ-DPG with $n\geq 2$ outperforms DDPG regarding control accuracy.}
  \label{fig:fig2}
 \end{figure*}

 Network architecture of hybrid actors-critics and other hyper-parameters are listed in  Appendix \ref{app:NN}. Note that we use an identical network architecture and learning algorithm hyper-parameters to different reference trajectories and different algorithms. Keeping exploring is another challenge in continuous action spaces. Fortunately, an advantage of off-policy algorithms such as our proposed MPQ-DPG is that the exploration can be handled independently from the learning process. And a common approach is to add noise sampled from a noise process $\mathcal{N}$. Here we use temporally correlated noise as DDPG for the purpose of exploring efficiently in trajectory tracking control problem of AUVs with inertia. And an Ornstein-Uhlenbeck process \cite{Uhlenbeck1945On} with $\theta=0.15$ and $\sigma=0.32$ is adopted for all simulations.

\subsection{Comparisons with DDPG}
 To verify the effectiveness of our proposed MPQ-DPG algorithm in the stability of learning and high-level tracking control accuracy when implemented to the trajectory tracking control of AUVs, we present a comparison between MPQ-DPG and DDPG. For the convenience of comparison, the number of critics is considered equal to that of actors in MPQ-DPG, namely, $n=m$. The results on both RT1 and RT2 are shown in Fig.
 \ref{fig:fig1}\text{--}\ref{fig:fig4} and Table. \ref{table:statistical}.
 \begin{table*}[!t]
  \renewcommand\arraystretch{1.5}
  \caption {Statistics of the total reward of DDPG and MPQ-DPG with $n=2,3,4,5,6$ on both RT1 and RT2.}
  \label{table:statistical}
  \begin{center}
  \scriptsize
  \begin{tabular}{|c|p{0.7cm}<{\centering}|c|c|c|c|c|c|c|c|c|}
   \hline
   \multicolumn{2}{|c|}{\multirow{2}{*}{Algorithms}}&\multicolumn{4}{c|}{RT1}&\multicolumn{4}{c|}{RT2}\\ \cline{3-10}
   \multicolumn{2}{|c|}{}            & $R_{best}$ & $R_{av}$ & $\text{STD-DEV}_R$ &
   $IR_n$ & $R_{best}$ & $R_{av}$ & $\text{STD-DEV}_R$ & $IR_n$ \\ \cline{1-10}
   \multicolumn{2}{|c|}{DDPG}        & -333.16 & -2500.8 & 4720.1 & $\verb|\|$ & -281.91 & -15746.1 & 44999.9 & $\verb|\|$  \\ \cline{1-10}
   \multirow{5}{*}{MPQ-DPG}  & $n=2$ & -207.29 & -416.11 & 326.471 & 0.83 & -187.93 & -601.33 & 2795.5 & 0.96 \\ \cline{2-10}
                             & $n=3$ & -192.53 & -362.95 & 319.65  & 0.86 & -141.47 & -244.21 & 166.30 & 0.98 \\
   \cline{2-10}
                             & $n=4$ & -192.46 & -330.26 & 96.45   & 0.87 & -137.52 & -267.49 & $\bm{70.09}$  & 0.98 \\
   \cline{2-10}
                             & $n=5$ & -133.94 & -311.14 & $\bm{68.92}$ & 0.88 & -117.95 & $\bm{-243.04}$ & 88.51 & $\bm{0.99}$ \\ \cline{2-10}
                             & $n=6$ & $\bm{-105.59}$ & $\bm{-282.52}$ & 83.22  & $\bm{0.89}$ & $\bm{-101.56}$ & -274.83 & 81.25  & 0.98 \\ \hline
   \end{tabular}
  \end{center}
 \end{table*}
 \begin{figure*}[!t]
  \centering
  \begin{minipage}[t]{0.5\linewidth}
   \centering
   \includegraphics*[width=0.82\linewidth,viewport=0 175 570 690]{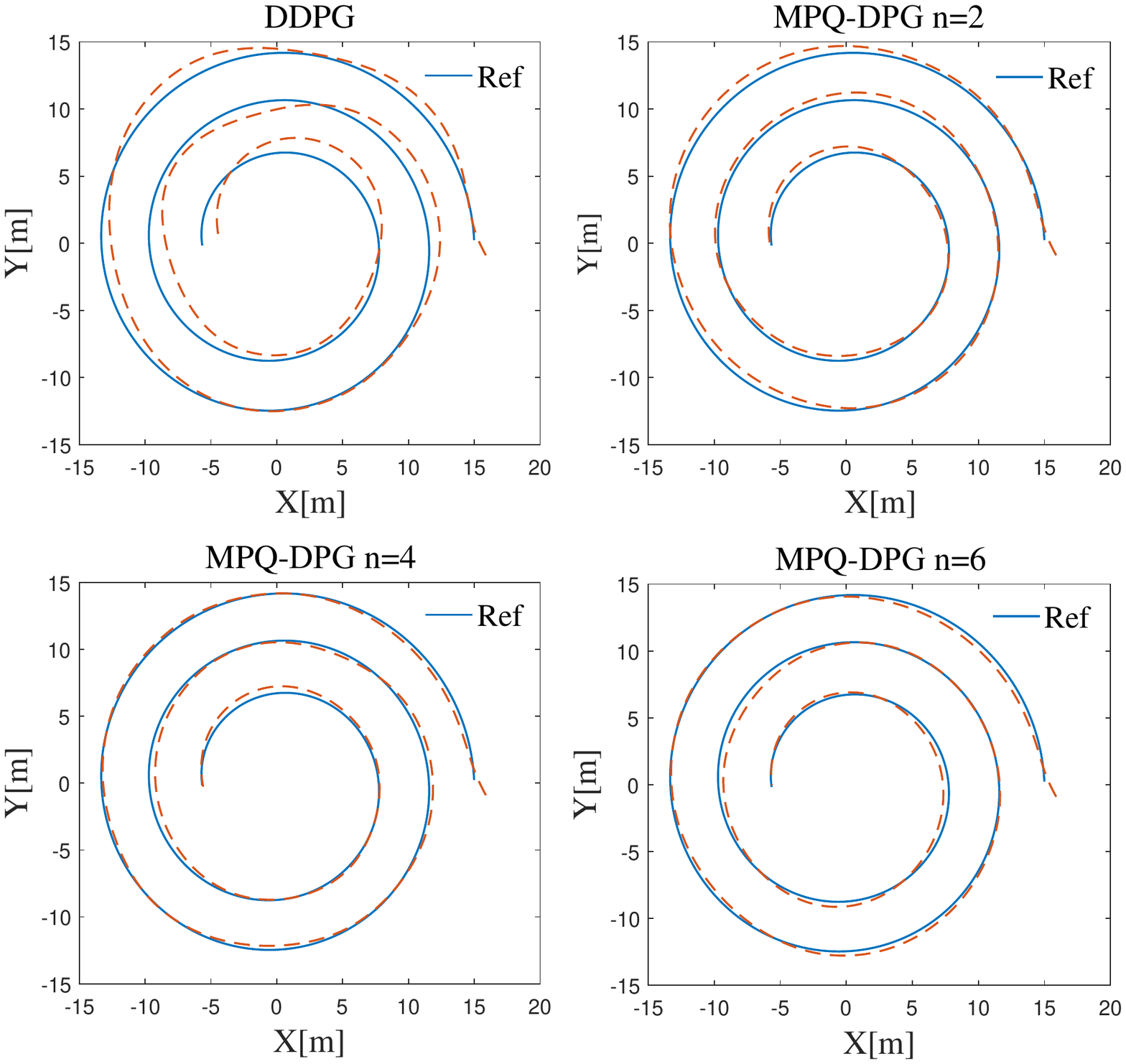}
  \end{minipage}%
  \begin{minipage}[t]{0.5\linewidth}
   \centering
   \includegraphics*[width=0.82\linewidth,viewport=0 175 570 690]{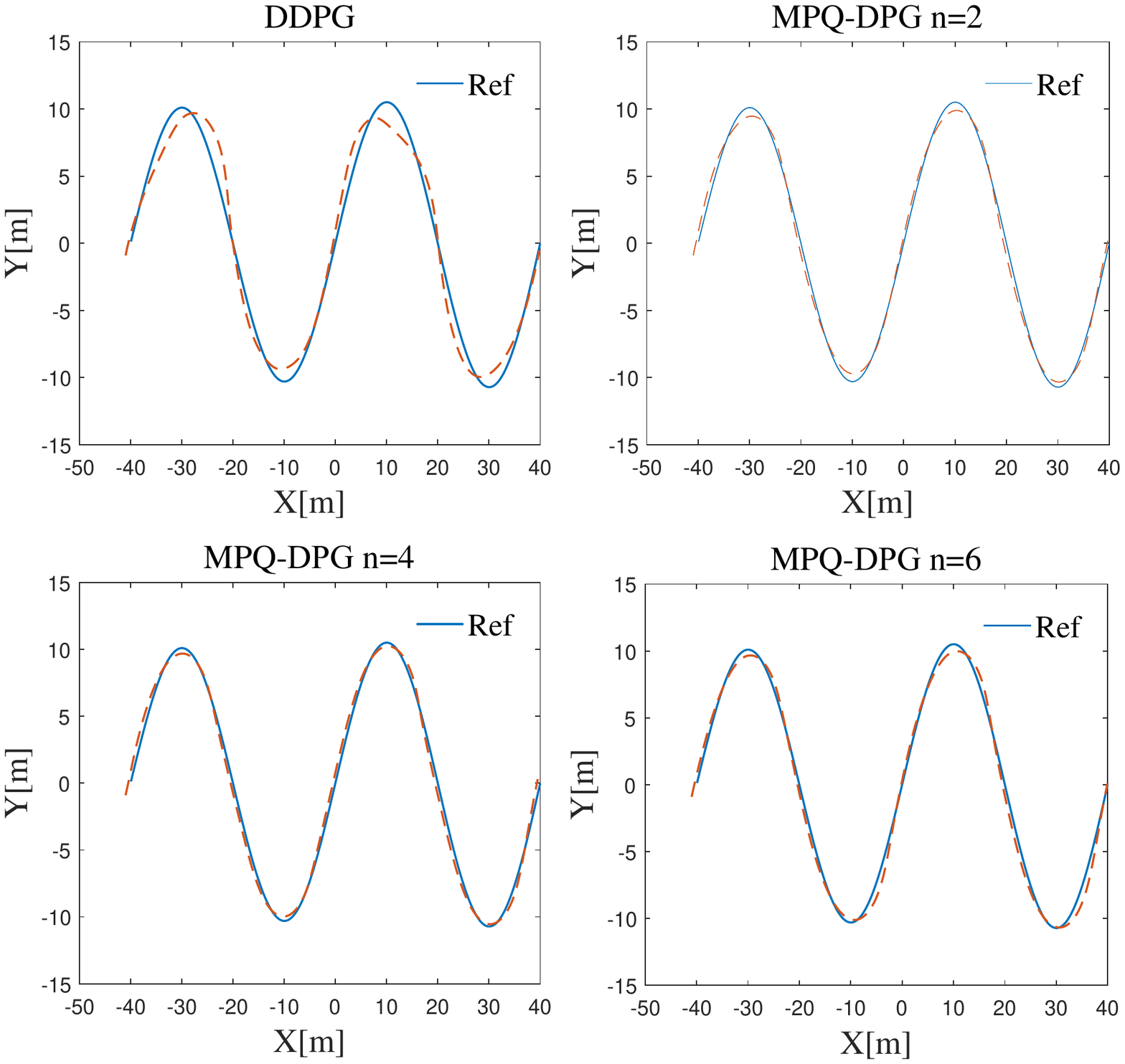}
  \end{minipage}
  \caption{Trajectories of REMUS obtained by DDPG and MPQ-DPG ($n=2,4,6$) with different reference trajectories. The left and right subfigures show the results corresponding to RT1 and RT2, respectively. For convenience to compare these algorithms, all trajectories are chosen as the best results in one learning trial. And the \textbf{red dashed} curves in all subfigures represent the trajectories obtained by simulations.}
  \label{fig:fig3}
 \end{figure*}

 The learning curves depicting the total reward per episode in one learning trial of each algorithm are shown in Fig. \ref{fig:fig1}. Although both MPQ-DPG ($n=2,3,4,5,6$) and DDPG converge to satisfactory values eventually, MPQ-DPG ($n=2,3,4,5,6$) performs better regarding the stability of learning process as was predicted by the theoretical analysis in Section \ref{sec:algorithm}. And these results provide empirical evidence that Multi Pseudo Q-learning (MPQ) is a more effective way to stabilize the learning, compared to the target Q network used in DDPG.

 Other simulation results provided in Fig. \ref{fig:fig2}--\ref{fig:fig3} demonstrate that MPQ-DPG ($n=2,3,4,5,6$) outperforms DDPG on both RT1 and RT2 regarding high-level accuracy of tracking control. Fig. \ref{fig:fig2} shows the average total reward over five independent learning trials, it can be seen that MPQ-DPG ($n=2,3,4,5,6$) is superior to DDPG regarding the average total reward maximization. And it is clear in Fig. \ref{fig:fig3} that MPQ-DPG ($n=2,4,6$) has smaller tracking error and obvious advantage in tracking sharp trajectory segments with large curvature when compared to DDPG. Furthermore, it can be observed in Fig. \ref{fig:fig2} that increasing the number of critics and actors tends to result in greater average total reward, but at the expense of slower learning speed. Actually, there exists a tradeoff between the performance and learning speed, but the performance difference between different number of critics ($n=2,3,4,5,6$) is small according to Fig. \ref{fig:fig2}--\ref{fig:fig3}. Hence, more attention should be paid to the learning speed when considering this tradeoff.

 In addition, Table. \ref{table:statistical} summarizes the principal statistical results. $R$ is a sequence of average total rewards over five independent learning trials corresponding to 1500 episodes. $R_{best}$ is the best or maximum value of $R$. $R_{av}$ and $\text{STD-DEV}_R$ are the mean and standard deviation of $R$ on episode interval [500,1500], respectively. And $IR_n$ represents the improvement rate defined as
 \begin{align}
  IR_n=1-\frac{R_{av}^{(n)}}{R_{av}^{(\text{DDPG})}}. \nonumber
 \end{align}
 As can be seen in Table. \ref{table:statistical}, for both RT1 and RT2, MPQ-DPG ($n=2,3,4,5,6$) shows better performance than DDPG on all statistics. Moreover, increasing the number of critics and actors $n$ in MPQ-DPG further leads to performance improvement, but which is no longer apparent when $n>4$,and the reason for this may be that 4 critics have been enough to reduce the overestimation and obtain an accurate estimation of target value. Actually, for other applications, the number of critics and actors can be chosen on the basis of three principles. First, overlarge number of critics are not suggested due to corresponding slow learning speed, in general, 2-4 critics are enough to satisfy the requirement of accuracy for most applications \cite{Duryea2016Exploring}. Second, according to Fig. \ref{fig:fig2} and Table. \ref{table:statistical}, increasing the number of critics will no longer obviously improve the performance after a particular number. Third, the number of critics should be suitably chosen to match the computing power of hardware.

 To validate the ability of EABE-based updating rule to accelerate the convergence of learning, we also report the results with EABE-based updating rule replaced by a stochastic updating rule for critics in MPQ-DPG ($n=2,3$). And Fig. \ref{fig:fig4} shows the performance curves with two different reference trajectories. It can be seen on both RT1 and RT2 that MPQ-DPG using EABE-based updating rule for critics obviously shortens the convergence time (i.e. the number of episodes required to converge) of the average total reward, compared to MPQ-DPG using a stochastic updating rule to determine which critic to be trained in each time step.
 \begin{figure}[!t]
  \centering
  \scriptsize
    \includegraphics*[width=0.8\linewidth,viewport=40 215 515 635]{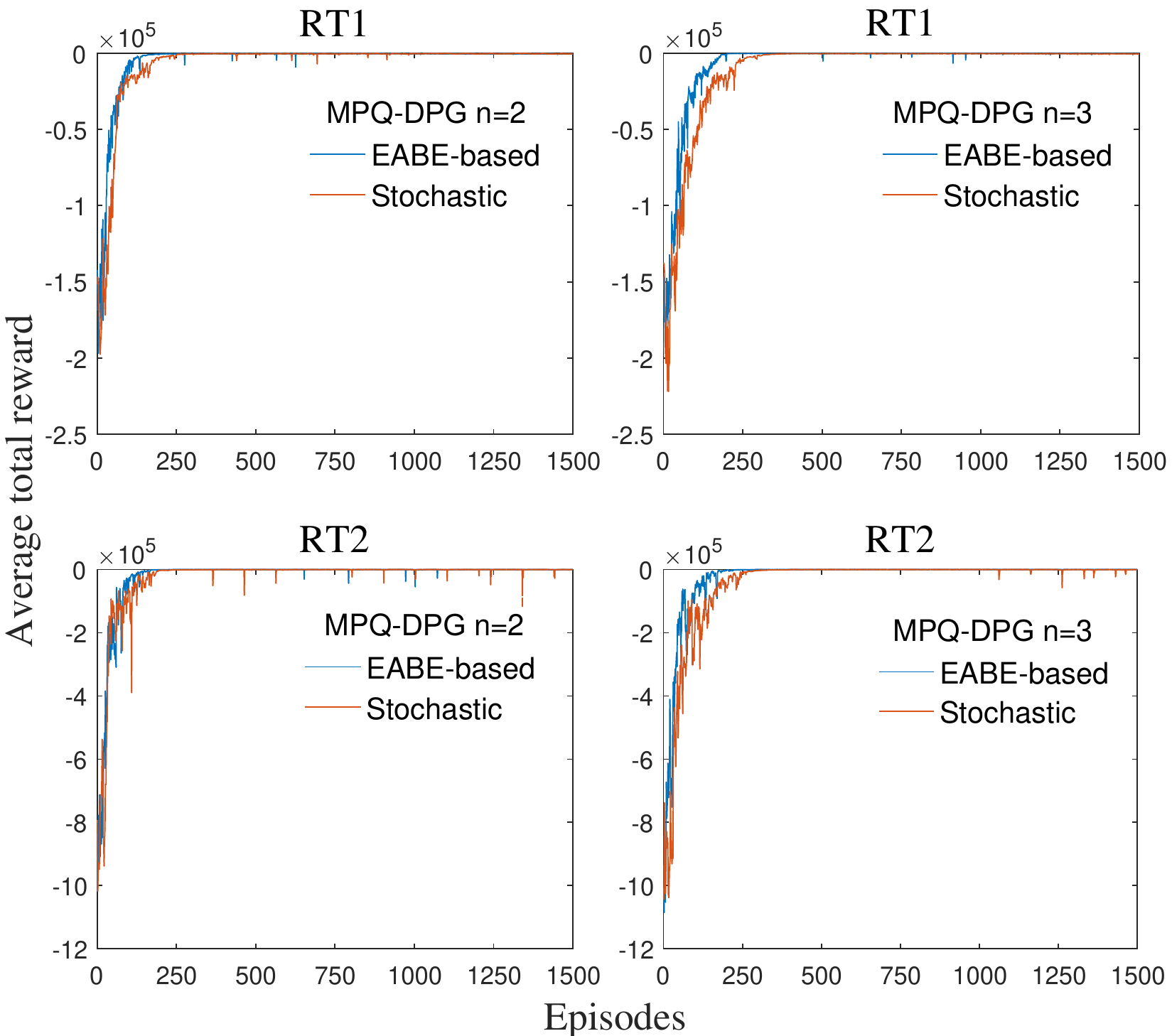}
  \caption{Performance (average total reward) curves of MPQ-DPG ($n=2,3$) with different updating rules for critics. All curves are averages over five independent learning trials. The \textbf{top} and \textbf{bottom} rows show the results corresponding to RT1 and RT2, respectively.}
  \label{fig:fig4}
 \end{figure}
 \begin{figure}[!t]
  \centering
  \scriptsize
    \includegraphics*[width=0.8\linewidth,viewport=70 242 525 585]{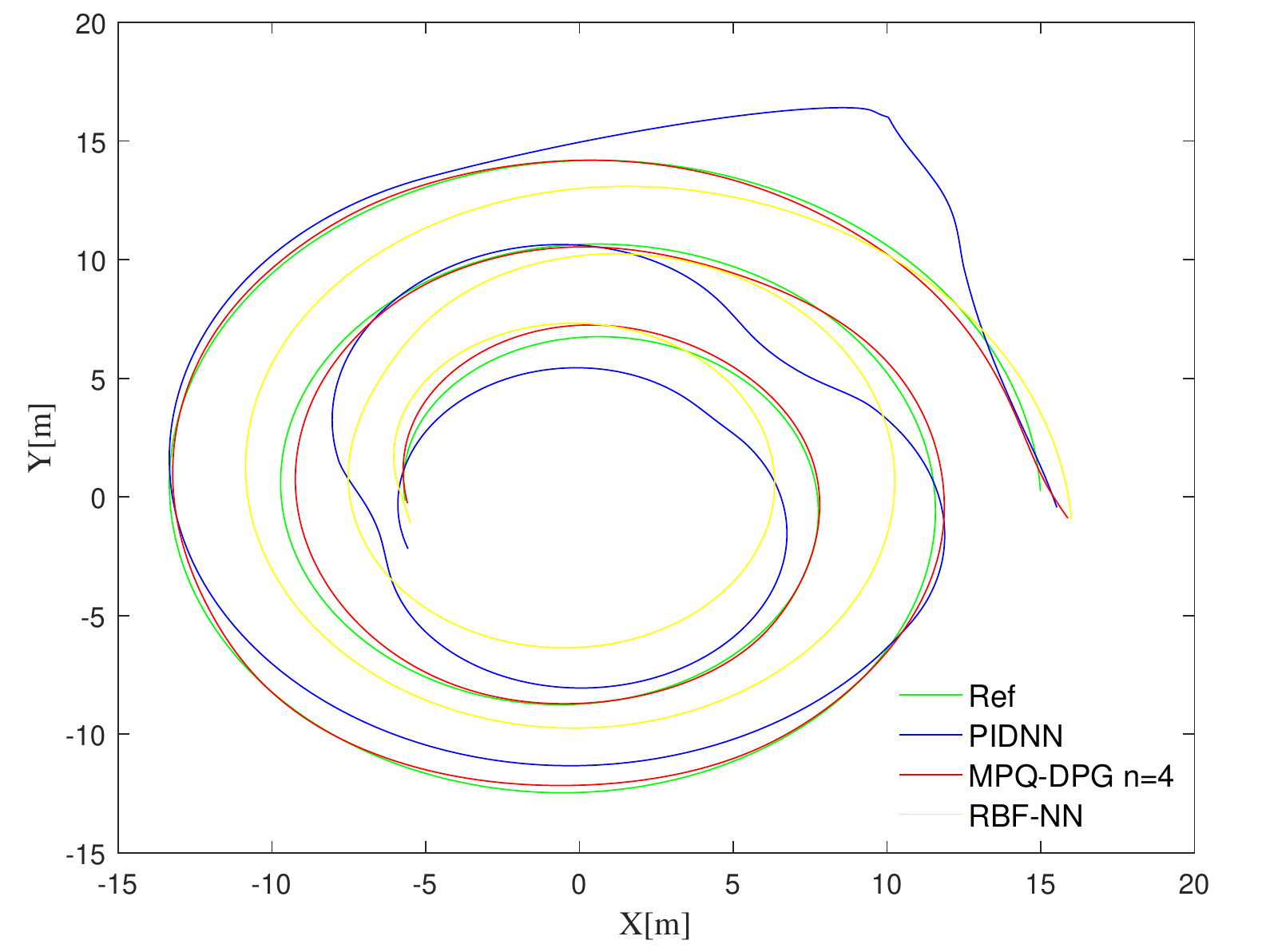}
  \caption{Trajectories of REMUS obtained by PIDNN control, RBF-NN control and MPQ-DPG ($n=4$).}
  \label{fig:fig7}
 \end{figure}

\subsection{Comparisons with PIDNN and RBF-NN Control}
 To further demonstrate the advantage of our proposed MPQ-DPG algorithm in term of high-level tracking control accuracy for AUVs, we compare MPQ-DPG ($n=4$) with PIDNN control and RBF-NN control on RT1. For PIDNN control and RBF-NN control, The initial states of REMUS are chosen as $\et=[16,-1,\pi/3]^{T}$, $\ph=[1.0,0.1,0]^{T}$, and the interval $T_s$ is selected as $0.02s$. Other hyper-parameters and network architecture of PIDNN control are listed in Appendix \ref{app:NN}. The results are presented in Fig. \ref{fig:fig7}\text{--}\ref{fig:fig5}.

\begin{figure}[!t]
  \centering
  \scriptsize
    \includegraphics*[width=0.8\linewidth,viewport=55 260 534 595]{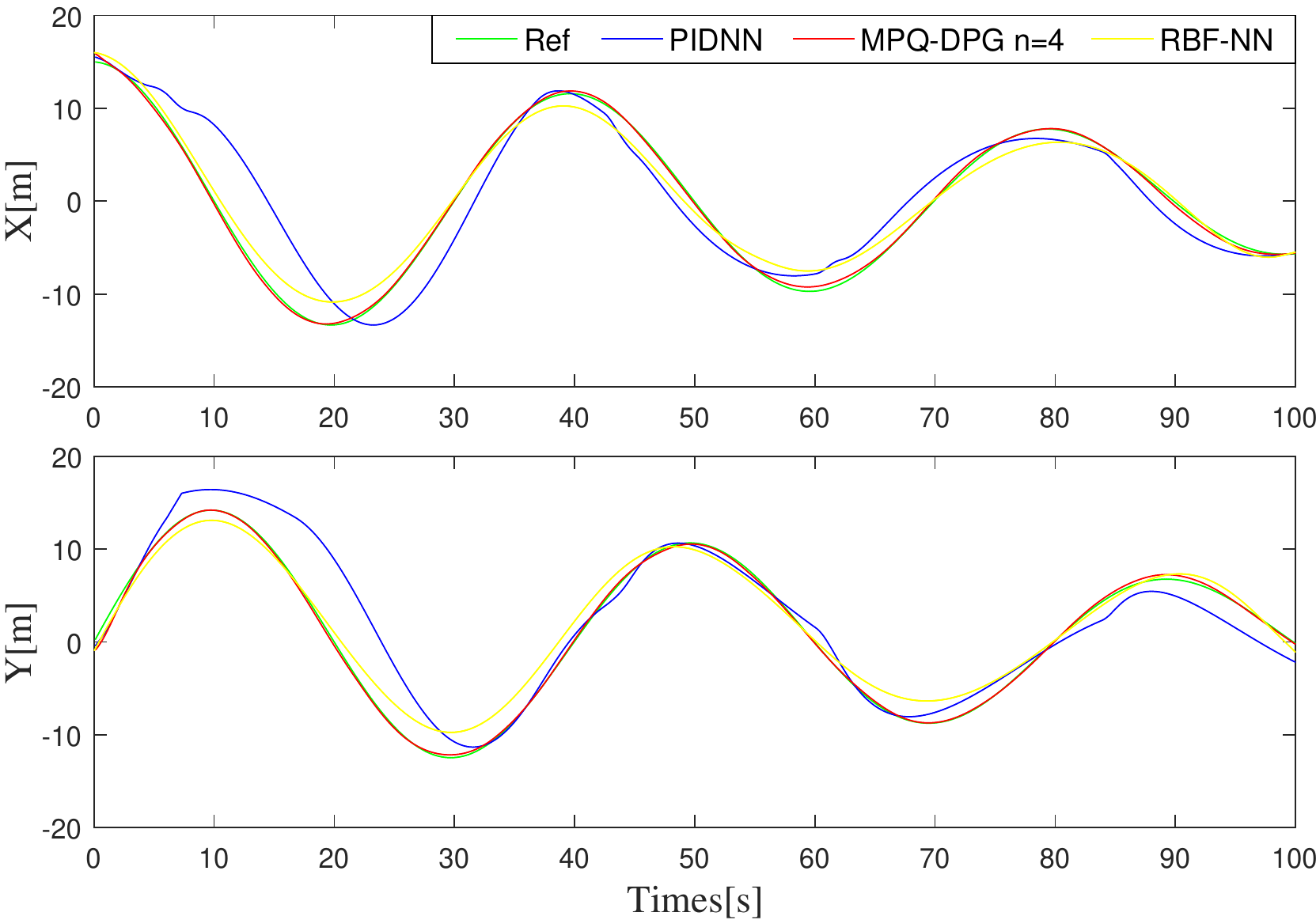}
  \caption{Tracking trajectories of coordinate positions of REMUS obtained by PIDNN control, RBF-NN control and MPQ-DPG ($n=4$).}
  \label{fig:fig8}
 \end{figure}

 It can be seen in Fig. \ref{fig:fig7}\text{--}\ref{fig:fig8} that generated trajectories can track the reference trajectory, the trajectories presented in Fig. \ref{fig:fig7} demonstrate high-level tracking control accuracy of MPQ-DPG, and the results in \ref{fig:fig8} show better performance of MPQ-DPG than that of PIDNN control and RBF-NN control when tracking sharp trajectory segments with large curvature. Moreover, we can conclude that MPQ-DPG obviously outperforms PIDNN control and RBF-NN control in term of tracking performance. And this partially attributes to powerful approximation ability of DNN.

\begin{figure}[!t]
  \centering
  \scriptsize
    \includegraphics*[width=0.8\linewidth,viewport=65 265 522 583]{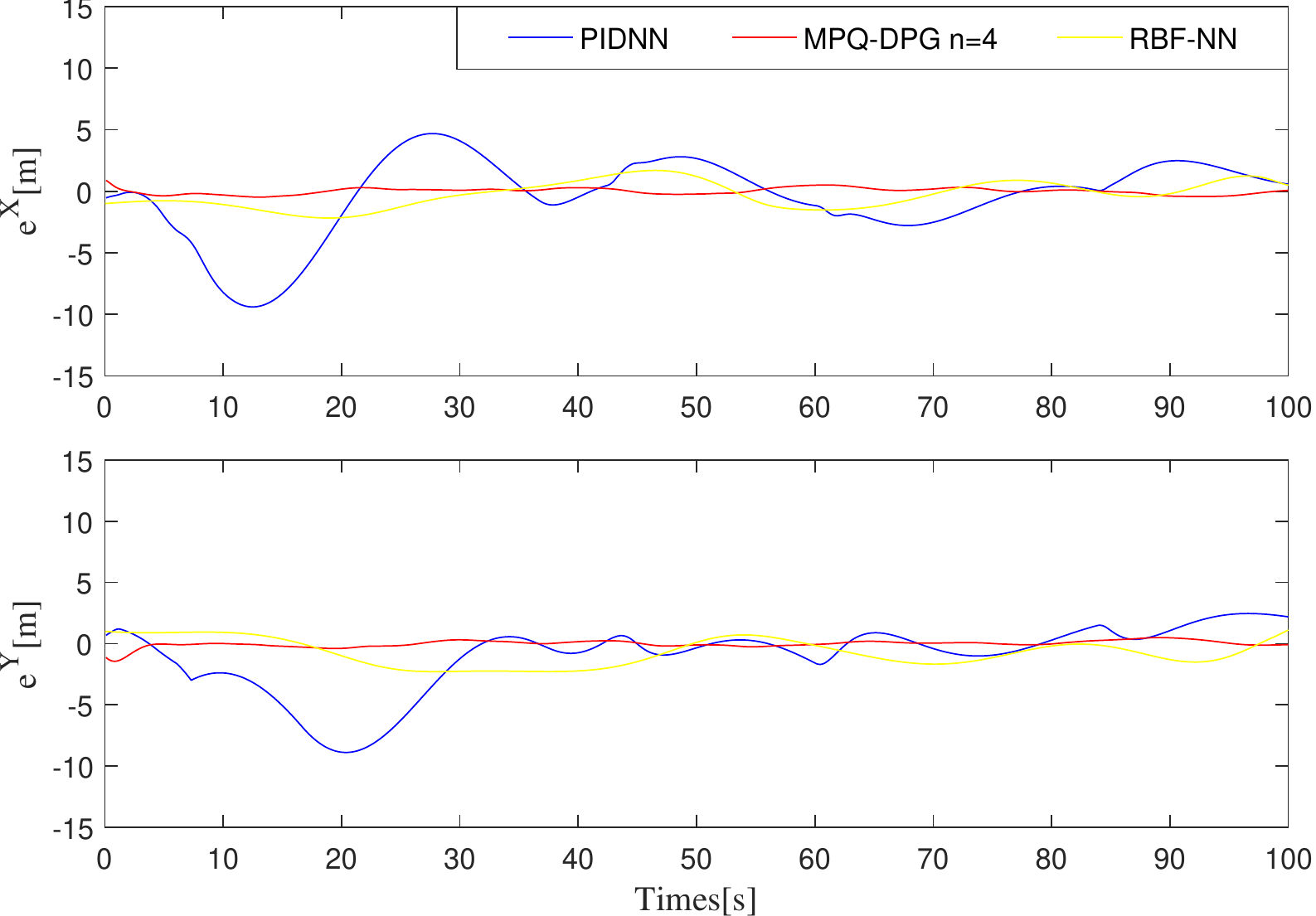}
  \caption{Tracking errors of REMUS obtained by PIDNN control, RBF-NN control and MPQ-DPG ($n=4$).}
  \label{fig:fig6}
 \end{figure}

\begin{figure}[!t]
  \centering
  \scriptsize
    \includegraphics*[width=0.8\linewidth,viewport=63 265 525 593]{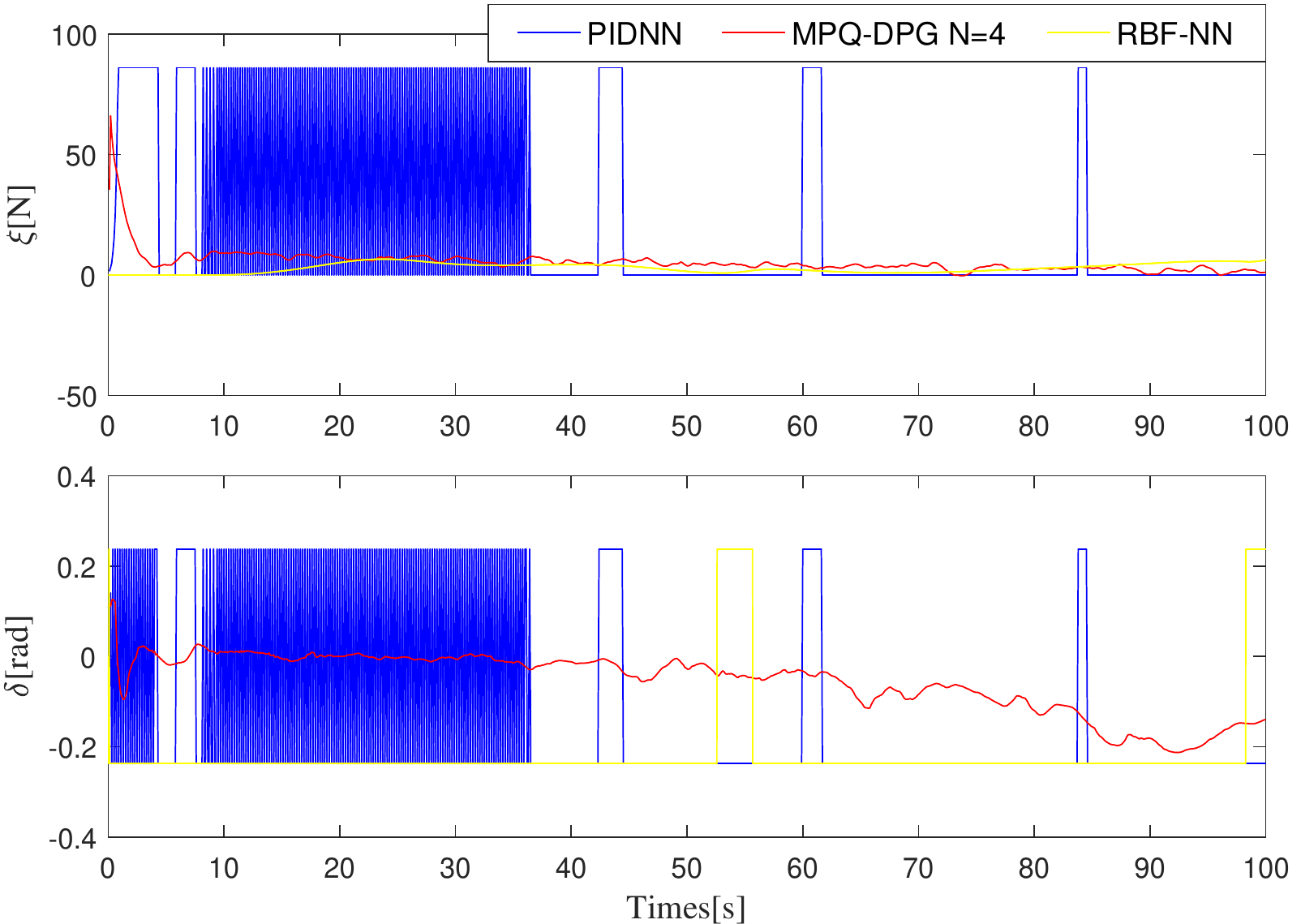}
  \caption{Control inputs of REMUS obtained by PIDNN control, RBF-NN control and MPQ-DPG ($n=4$). The top and bottom subfigures represent the propeller thrust and rudder angle, respectively.}
  \label{fig:fig5}
 \end{figure}

 The tracking errors are shown in Fig. \ref{fig:fig6}, it can be seen that MPQ-DPG achieves faster convergence of coordinate positions and smaller tracking error than PIDNN control or RBF-NN control. Moreover, Fig. \ref{fig:fig5} shows that the tracking controller found by MPQ-DPG with $n=4$ has smoother control input curves than that found by PIDNN control or RBF-NN control, consequently, we can conclude that the controller found by MPQ-DPG is more easy to operate in real applications.

 As can be seen in Fig. \ref{fig:fig7}\text{--}\ref{fig:fig5}, the tracking performance of PIDNN control is not satisfactory, although we have run the simulation many times to expect to choose the best initial weights for the neural network. And the reason for this is, on the one hand, attributed to the high nonlinearity and strong coupling of the underactuated AUVs system, and on the other hand due to the inherent disadvantage of this method in choosing appropriate hyper-parameters.

\section{Conclusion}\label{sec:conclusion}
 This paper presents a model-free policy gradient algorithm for the trajectory tracking problem of underactuated AUV with unknown dynamics and constrained inputs based on a hybrid actors-critics architecture, in which multiple actors and critics are trained to learn a deterministic policy and action-value function, respectively. And the final learned policy, defined as the average of all actors to avoid large but bad updates of policy, showed outstanding tracking performance.

 For multiple critics, the EABE-based updating rule was developed to determine which critic to update in each time step, and the simulation results have shown that this updating rule can accelerate the convergence of learning by weakening the effect of bad critics. Moreover, as an extension of Q-learning to RL problems with continuous action spaces, Pseudo Q-learning, which adopted the sub-greedy policy to replace the greedy policy in Q-learning, has shown the advantage to calculate the loss function with more accurate target value when updating the critic. Meanwhile, Multi Pseudo Q-learning (MPQ) was proposed to reduce the overestimation resulting from the maximization in sub-greedy policy and to stabilize the learning. The simulation results provided empirical evidence that MPQ-DPG outperforms DDPG and PIDNN in terms of high-level tracking control accuracy for AUVs and the stability of learning. The results also demonstrated that MPQ is more efficient to stabilize the learning, compared to the target Q network used in DDPG.

 Future work will explore ways to further accelerate the learning and focus on conducting experiments on real-world AUVs. A possible approach is to start training with single critic, and then gradually increase the number of critics by using the weights been updated several episodes ago to initialize the new critics. In addition, transfer learning on the real AUVs can further improve the performance of our proposed algorithm by capturing unknown dynamic aspects of environment.

\begin{appendices}
\section{Model matrices and Coefficients of REMUS} \label{app:model coefficients}
 The model matrices $\mb$, $\cb(\ph)$, $\db(\ph)$ and $\gb(\ph)$ of REMUS are given by
 \begin{gather}
  M=\begin{bmatrix} \label{eqn:M}
      m_1 & 0   & 0    \\
      0   & m_2 & m_3  \\
      0   & m_4 & m_5
    \end{bmatrix}. \nonumber
 \end{gather}
 \begin{gather} \label{eqn:C}
  C(\phi)=\begin{bmatrix}
      0         & 0       & c_1(v,r)   \\
      0         & 0       & c_2(u)     \\
      -c_1(v,r) & -c_2(u) & 0
    \end{bmatrix}. \nonumber
 \end{gather}
 \begin{gather}
  D(\phi)=\begin{bmatrix} \label{eqn:D}
      d_1(u)  & 0        & 0         \\
      0       & d_2(u,v) & d_3(u,r)  \\
      0       & d_4(u,v) & d_5(u,r)
    \end{bmatrix}. \nonumber
 \end{gather}
 \begin{gather} \label{eqn:G}
  G(\phi)=\begin{bmatrix}
      1       & 0                 \\
      0       & Y_{uu\delta}u^2   \\
      0       & N_{uu\delta}u^2
    \end{bmatrix}. \nonumber
 \end{gather}
 where $m_1=m-X_{\dot{u}},~m_2=m-Y_{\dot{v}},~m_3=mx_g-Y_{\dot{r}},~m_4=mx_g-N_{\dot{v}},~m_5=I_{zz}-N_{\dot{r}},~c_1=-mv-mx_gr+ Y_{\dot{v}}v+Y_{\dot{r}}r,~c_2=mu-X_{\dot{u}}u,~d_1=-X_u-X_{u|u|}|u|,~d_2=-Y_v-Y_{uv}u-Y_{v|v|}|v|,~d_3=-Y_r-Y_{ur}u-Y_{rr}|r|,~ d_4=-N_v-N_{uv}u-N_{v|v|}|v|,~d_5=-N_r-N_{ur}u-N_{rr}|r|$, $m$ is the mass of AUV, $x_g$ is the $x$-position of the center of gravity, $I_{zz}$ is the mass moment of inertia term, and other coefficients are hydrodynamic coefficients explained in \cite{fossen2011handbook}.

 For all simulations in this paper, the coefficients of REMUS are adopted as $m=3.048\times 10~\text{kg}$, $x_g=0$, $I_zz=3.45~\text{kg}\cdot \text{m}^2$, $X_u=0$, $X_{u|u|}=-1.62~\text{kg}/\text{m}$, $X_{\dot{u}}=-9.30\times 10^{-1}\text{kg}$, $Y_v=0$, $Y_r=0$, $Y_{v|v|}=-1.31\times 10^3~\text{kg}/\text{m}$, $Y_{rr}=6.32\times 10^{-1}~\text{kg}\cdot\text{m}/\text{rad}^2$, $Y_{uv}=-2.86\times 10~\text{kg}/\text{m}$, $Y_{\dot{v}}=-3.55\times 10~\text{kg}$, $Y_{\dot{r}}=1.93~\text{kg}\cdot\text{m}/\text{rad}$, $Y_{ur}=6.15~\text{kg}/\text{rad}$, $Y_{uu\delta}=9.64~\text{kg}/(\text{m}\cdot\text{rad})$, $N_v=0$, $N_r=0$, $N_{v|v|}=-3.18~\text{kg}$, $N_{rr}=-9.40\times 10~\text{kg}\cdot\text{m}^2/\text{rad}^2$, $N_{uv}=10.62~\text{kg}$, $N_{\dot{v}}=1.93\text{kg}\cdot\text{m}$, $N_{\dot{r}}=-4.88~\text{kg}\cdot\text{m}^2/\text{rad}$, $N_{ur}=-3.93~\text{kg}\cdot\text{m}/\text{rad}$, and $N_{uu\delta}=-6.15~\text{kg}/\text{rad}$.

\section{Network Architecture and Hyper-Parameters} \label{app:NN}
 We used Adam \cite{Kingma2014Adam} for learning the neural network parameters with a learning rate of $10^{-4}$ and $10^{-3}$ for the actor and critic, respectively. For action-value function $Q$, we included $L_2$ weight decay of $10^{-2}$ and used a discounted factor of $\gamma=0.99$. The neural networks used the Rectified non-Linearity Unit (ReLu) \cite{glorot2011deep} for all hidden layers. The final output layer of the actor was a tanh layer to bound the actions. The low-dimensional actor-critic networks both had 2 hidden layers with 400 and 300 units, respectively. And for critic networks, actions were not included until the 2nd hidden layer of Q. The final layer weights and bias of both the actors and critics were initialized from a uniform distribution $[-3\times 10^{-3},3\times 10^{-3}]$ to ensure the initial outputs for the policy and action-value estimates were near zero. The other layers were all initialized from uniform distributions $[-\frac{1}{\sqrt{f}},\frac{1}{\sqrt{f}}]$ where $f$ is the fan-in of the current layer. In all experiments, training was done over 1500 episodes and each episode included 1000 time steps. The size of the experience replay buffer is 10000 tuples. The buffer got sampled to updated critics and actors every time step with minibatches of size 64. Moreover, all MDP states were normalized with $[-1,1]$ to guarantee a unified scale.

 For the PIDNN controller, we applied a common network structure with three layers consisting of an input layer, a hidden layer and an output layer. The hidden layer comprised two group of PID neurons corresponding to two control states, and each group of PID neurons included a proportion neuron, an integral neuron and a differential neuron. The hidden layer was fully connected with output layer. The weights were all initialized by particle swarm optimization (PSO) algorithm. The learning rates of two group of weights were 0.005 and 0.001, respectively.

\end{appendices}


\bibliography{reference}

\begin{thebibliography}{10}
\providecommand{\url}[1]{#1}
\csname url@samestyle\endcsname
\providecommand{\newblock}{\relax}
\providecommand{\bibinfo}[2]{#2}
\providecommand{\BIBentrySTDinterwordspacing}{\spaceskip=0pt\relax}
\providecommand{\BIBentryALTinterwordstretchfactor}{4}
\providecommand{\BIBentryALTinterwordspacing}{\spaceskip=\fontdimen2\font plus
\BIBentryALTinterwordstretchfactor\fontdimen3\font minus
  \fontdimen4\font\relax}
\providecommand{\BIBforeignlanguage}[2]{{%
\expandafter\ifx\csname l@#1\endcsname\relax
\typeout{** WARNING: IEEEtran.bst: No hyphenation pattern has been}%
\typeout{** loaded for the language `#1'. Using the pattern for}%
\typeout{** the default language instead.}%
\else
\language=\csname l@#1\endcsname
\fi
#2}}
\providecommand{\BIBdecl}{\relax}
\BIBdecl

\bibitem{Singh2011Control}
M.~P. Singh and B.~Chowdhury, ``Control of autonomous underwater vehicles,''
  2011.

\bibitem{das2016cooperative}
B.~Das, B.~Subudhi, and B.~B. Pati, ``Cooperative formation control of
  autonomous underwater vehicles: An overview,'' \emph{International Journal of
  Automation and Computing}, vol.~13, no.~3, pp. 199--225, 2016.

\bibitem{refsnes2008model}
J.~E. Refsnes, A.~J. Sorensen, and K.~Y. Pettersen, ``Model-based output
  feedback control of slender-body underactuated auvs: Theory and
  experiments,'' \emph{IEEE Transactions on control systems technology},
  vol.~16, no.~5, pp. 930--946, 2008.

\bibitem{li2016trajectory}
Z.~Li, J.~Deng, R.~Lu, Y.~Xu, J.~Bai, and C.-Y. Su, ``Trajectory-tracking
  control of mobile robot systems incorporating neural-dynamic optimized model
  predictive approach,'' \emph{IEEE Transactions on Systems, Man, and
  Cybernetics: Systems}, vol.~46, no.~6, pp. 740--749, 2016.

\bibitem{kim2015tracking}
D.~W. Kim, ``Tracking of remus autonomous underwater vehicles with actuator
  saturations,'' \emph{Automatica (Journal of IFAC)}, vol.~58, no.~C, pp.
  15--21, 2015.

\bibitem{zhu2013dynamic}
D.~Zhu, H.~Huang, and S.~X. Yang, ``Dynamic task assignment and path planning
  of multi-auv system based on an improved self-organizing map and velocity
  synthesis method in three-dimensional underwater workspace,'' \emph{IEEE
  Transactions on Cybernetics}, vol.~43, no.~2, pp. 504--514, 2013.

\bibitem{cui2016mutual}
R.~Cui, Y.~Li, and W.~Yan, ``Mutual information-based multi-auv path planning
  for scalar field sampling using multidimensional rrt,'' \emph{IEEE
  Transactions on Systems, Man, and Cybernetics: Systems}, vol.~46, no.~7, pp.
  993--1004, 2016.

\bibitem{peng2017output}
Z.~Peng and J.~Wang, ``Output-feedback path-following control of autonomous
  underwater vehicles based on an extended state observer and projection neural
  networks,'' \emph{IEEE Transactions on Systems, Man, and Cybernetics:
  Systems}, 2017.

\bibitem{xiang2016path}
X.~Xiang, C.~Yu, Q.~Zhang, and G.~Xu, ``Path-following control of an auv: Fully
  actuated versus under-actuated configuration,'' \emph{Marine Technology
  Society Journal}, vol.~50, no.~1, pp. 34--47, 2016.

\bibitem{healey1993multivariable}
A.~J. Healey and D.~Lienard, ``Multivariable sliding mode control for
  autonomous diving and steering of unmanned underwater vehicles,'' \emph{IEEE
  journal of Oceanic Engineering}, vol.~18, no.~3, pp. 327--339, 1993.

\bibitem{Subudhi2013A}
B.~Subudhi, K.~Mukherjee, and S.~Ghosh, ``A static output feedback control
  design for path following of autonomous underwater vehicle in vertical
  plane,'' \emph{Ocean Engineering}, vol.~63, no.~5, pp. 72--76, 2013.

\bibitem{He2015Vibration}
W.~He and S.~S. Ge, ``Vibration control of a flexible beam with output
  constraint,'' \emph{IEEE Transactions on Industrial Electronics}, vol.~62,
  no.~8, pp. 5023--5030, 2015.

\bibitem{cui2010leader}
R.~Cui, S.~S. Ge, B.~V.~E. How, and Y.~S. Choo, ``Leader--follower formation
  control of underactuated autonomous underwater vehicles,'' \emph{Ocean
  Engineering}, vol.~37, no.~17, pp. 1491--1502, 2010.

\bibitem{wang2015neural}
H.~Wang, K.~Liu, X.~Liu, B.~Chen, and C.~Lin, ``Neural-based adaptive
  output-feedback control for a class of nonstrict-feedback stochastic
  nonlinear systems,'' \emph{IEEE transactions on cybernetics}, vol.~45, no.~9,
  pp. 1977--1987, 2015.

\bibitem{he2016adaptive1}
W.~He, Y.~Chen, and Y.~Zhao, ``Adaptive neural network control of an uncertain
  robot with full-state constraints,'' \emph{IEEE Transactions on Cybernetics},
  vol.~46, no.~3, pp. 620--629, 2016.

\bibitem{he2016adaptive2}
W.~He, Y.~Dong, and C.~Sun, ``Adaptive neural impedance control of a robotic
  manipulator with input saturation,'' \emph{IEEE Transactions on Systems, Man,
  and Cybernetics: Systems}, vol.~46, no.~3, pp. 334--344, 2016.

\bibitem{sun2017adaptive}
C.~Sun, W.~He, W.~Ge, and C.~Chang, ``Adaptive neural network control of biped
  robots,'' \emph{IEEE Transactions on Systems, Man, and Cybernetics: Systems},
  vol.~PP, no.~99, pp. 1--12, 2017.

\bibitem{wei2017adaptive}
W.~He, Y.~Zhao, and C.~Sun, ``Adaptive neural network control of a marine
  vessel with constraints using the asymmetric barrier lyapunov function,''
  \emph{IEEE Transactions on Cybernetics}, vol.~47, no.~7, pp. 1641--1651,
  2017.

\bibitem{Konar2013ADI}
A.~Konar, I.~Goswami, S.~J. Singh, L.~C. Jain, and A.~K. Nagar, ``A
  deterministic improved q-learning for path planning of a mobile robot,''
  \emph{IEEE Trans. Systems, Man, and Cybernetics: Systems}, vol.~43, pp.
  1141--1153, 2013.

\bibitem{song2015multiple}
R.~Song, F.~Lewis, Q.~Wei, H.-G. Zhang, Z.-P. Jiang, and D.~Levine, ``Multiple
  actor-critic structures for continuous-time optimal control using
  input-output data,'' \emph{IEEE transactions on neural networks and learning
  systems}, vol.~26, no.~4, pp. 851--865, 2015.

\bibitem{kamalapurkar2017model}
R.~Kamalapurkar, L.~Andrews, P.~Walters, and W.~E. Dixon, ``Model-based
  reinforcement learning for infinite-horizon approximate optimal tracking,''
  \emph{IEEE transactions on neural networks and learning systems}, vol.~28,
  no.~3, pp. 753--758, 2017.

\bibitem{mu2017air}
C.~Mu, Z.~Ni, C.~Sun, and H.~He, ``Air-breathing hypersonic vehicle tracking
  control based on adaptive dynamic programming,'' \emph{IEEE transactions on
  neural networks and learning systems}, vol.~28, no.~3, pp. 584--598, 2017.

\bibitem{ouyang2017reinforcement}
Y.~Ouyang, W.~He, and X.~Li, ``Reinforcement learning control of a single-link
  flexible robotic manipulator,'' \emph{IET Control Theory \& Applications},
  vol.~11, no.~9, pp. 1426--1433, 2017.

\bibitem{gaskett1999reinforcement}
C.~Gaskett, D.~Wettergreen, A.~Zelinsky \emph{et~al.}, ``Reinforcement learning
  applied to the control of an autonomous underwater vehicle,'' in
  \emph{Proceedings of the Australian Conference on Robotics and Automation
  (AuCRA99)}, 1999.

\bibitem{carreras2005behavior}
M.~Carreras, J.~Yuh, J.~Batlle, and P.~Ridao, ``A behavior-based scheme using
  reinforcement learning for autonomous underwater vehicles,'' \emph{IEEE
  Journal of Oceanic Engineering}, vol.~30, no.~2, pp. 416--427, 2005.

\bibitem{el2013two}
A.~El-Fakdi and M.~Carreras, ``Two-step gradient-based reinforcement learning
  for underwater robotics behavior learning,'' \emph{Robotics and Autonomous
  Systems}, vol.~61, no.~3, pp. 271--282, 2013.

\bibitem{de2015trajectory}
M.~De~Paula and G.~G. Acosta, ``Trajectory tracking algorithm for autonomous
  vehicles using adaptive reinforcement learning,'' in \emph{OCEANS'15 MTS/IEEE
  Washington}.\hskip 1em plus 0.5em minus 0.4em\relax IEEE, 2015, pp. 1--8.

\bibitem{tai2017virtual}
L.~Tai, G.~Paolo, and M.~Liu, ``Virtual-to-real deep reinforcement learning:
  Continuous control of mobile robots for mapless navigation,'' \emph{arXiv
  preprint arXiv:1703.00420}, 2017.

\bibitem{huang2017parameterized}
Z.~Huang, X.~Xu, H.~He, J.~Tan, and Z.~Sun, ``Parameterized batch reinforcement
  learning for longitudinal control of autonomous land vehicles,'' \emph{IEEE
  Transactions on Systems, Man, and Cybernetics: Systems}, 2017.

\bibitem{hwangbo2017control}
J.~Hwangbo, I.~Sa, R.~Siegwart, and M.~Hutter, ``Control of a quadrotor with
  reinforcement learning,'' \emph{IEEE Robotics and Automation Letters},
  vol.~2, no.~4, pp. 2096--2103, 2017.

\bibitem{mannucci2017safe}
T.~Mannucci, E.-J. van Kampen, C.~de~Visser, and Q.~Chu, ``Safe exploration
  algorithms for reinforcement learning controllers,'' \emph{IEEE Transactions
  on Neural Networks and Learning Systems}, 2017.

\bibitem{sallab2017deep}
A.~E. Sallab, M.~Abdou, E.~Perot, and S.~Yogamani, ``Deep reinforcement
  learning framework for autonomous driving,'' \emph{Electronic Imaging}, vol.
  2017, no.~19, pp. 70--76, 2017.

\bibitem{mnih2015human}
V.~Mnih, K.~Kavukcuoglu, D.~Silver, A.~A. Rusu, J.~Veness, M.~G. Bellemare,
  A.~Graves, M.~Riedmiller, A.~K. Fidjeland, G.~Ostrovski \emph{et~al.},
  ``Human-level control through deep reinforcement learning,'' \emph{Nature},
  vol. 518, no. 7540, pp. 529--533, 2015.

\bibitem{sun2015target}
T.~Sun, B.~He, R.~Nian, and T.~Yan, ``Target following for an autonomous
  underwater vehicle using regularized elm-based reinforcement learning,'' in
  \emph{OCEANS'15 MTS/IEEE Washington}.\hskip 1em plus 0.5em minus 0.4em\relax
  IEEE, 2015, pp. 1--5.

\bibitem{silver2014deterministic}
D.~Silver, G.~Lever, N.~Heess, T.~Degris, D.~Wierstra, and M.~Riedmiller,
  ``Deterministic policy gradient algorithms,'' in \emph{Proceedings of the
  31st International Conference on Machine Learning (ICML-14)}, 2014, pp.
  387--395.

\bibitem{lillicrap2015continuous}
T.~P. Lillicrap, J.~J. Hunt, A.~Pritzel, N.~Heess, T.~Erez, Y.~Tassa,
  D.~Silver, and D.~Wierstra, ``Continuous control with deep reinforcement
  learning,'' \emph{arXiv preprint arXiv:1509.02971}, 2015.

\bibitem{sutton2011reinforcement}
R.~S. Sutton and A.~G. Barto, \emph{Reinforcement learning: An
  introduction}.\hskip 1em plus 0.5em minus 0.4em\relax Cambridge, MA: MIT
  Press, 2011.

\bibitem{van2016deep}
H.~Van~Hasselt, A.~Guez, and D.~Silver, ``Deep reinforcement learning with
  double q-learning.'' in \emph{AAAI}, 2016, pp. 2094--2100.

\bibitem{fossen2011handbook}
T.~I. Fossen, \emph{Handbook of marine craft hydrodynamics and motion
  control}.\hskip 1em plus 0.5em minus 0.4em\relax John Wiley \& Sons, 2011.

\bibitem{prestero2001verification}
T.~T.~J. Prestero, ``Verification of a six-degree of freedom simulation model
  for the remus autonomous underwater vehicle,'' Ph.D. dissertation,
  Massachusetts institute of technology, 2001.

\bibitem{Uhlenbeck1945On}
G.~E. Uhlenbeck and L.~S. Ornstein, ``On the theory of the brownian motion,''
  \emph{Physical Review}, vol.~17, no. 2-3, pp. 323--342, 1945.

\bibitem{Duryea2016Exploring}
E.~Duryea, M.~Ganger, and W.~Hu, ``Exploring deep reinforcement learning with
  multi q-learning,'' \emph{Intelligent Control and Automation}, vol.~07,
  no.~4, pp. 129--144, 2016.

\bibitem{Kingma2014Adam}
D.~P. Kingma and J.~Ba, ``Adam: A method for stochastic optimization,''
  \emph{Computer Science}, 2014.

\bibitem{glorot2011deep}
X.~Glorot, A.~Bordes, and Y.~Bengio, ``Deep sparse rectifier neural networks,''
  in \emph{Proceedings of the Fourteenth International Conference on Artificial
  Intelligence and Statistics}, 2011, pp. 315--323.

\end{thebibliography}

\bibliographystyle{IEEEtran}

\begin{IEEEbiography}[{\includegraphics[width=1in,height=1.25in,clip,keepaspectratio]{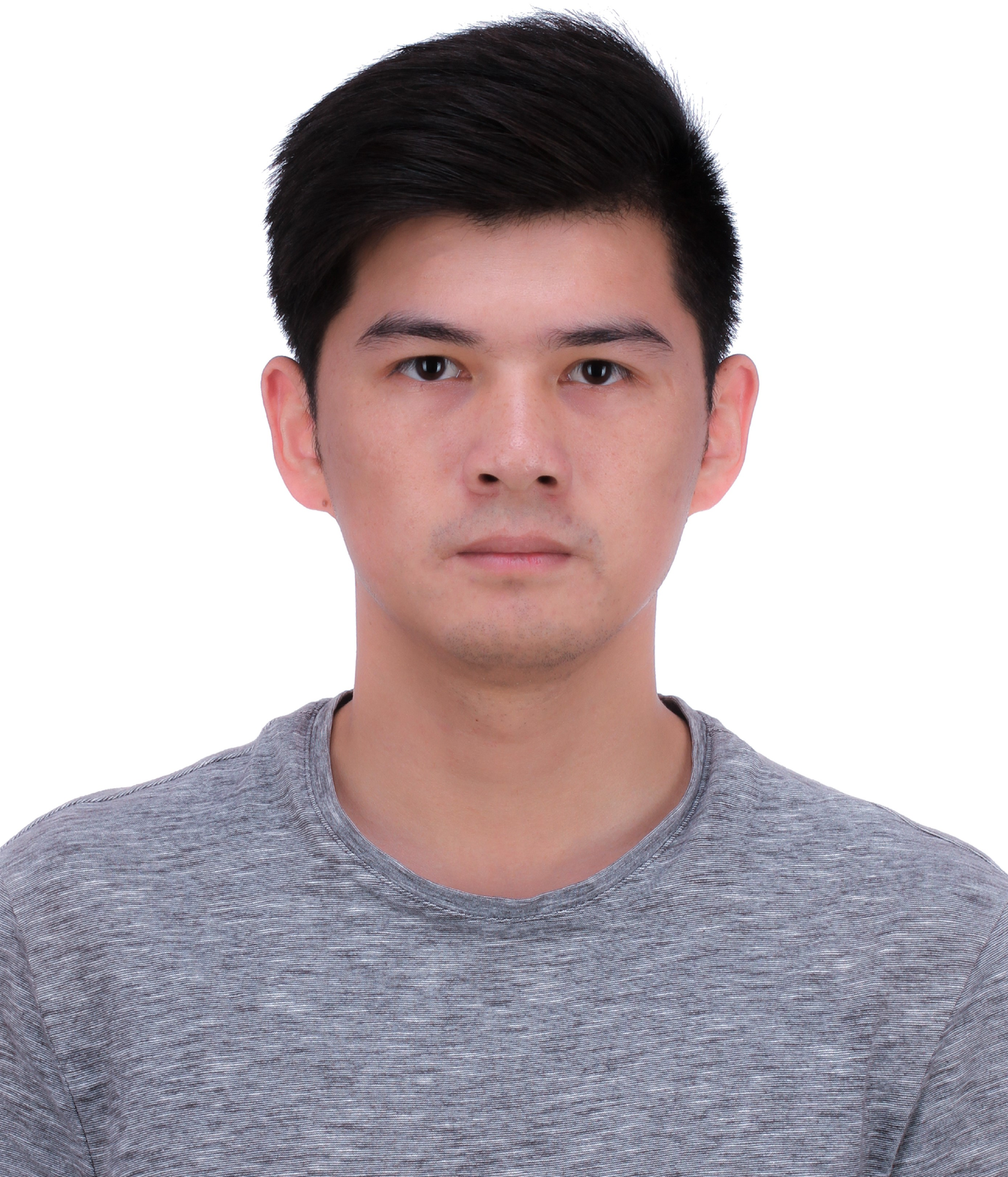}}]{Wenjie Shi}received the B.S. degree from the School of Hydropower and Information Engineering, Huazhong University of Science and Technology, Wuhan, China, in 2016. And he is currently pursuing the Ph.D. degree in control science and engineering with the Department of Automation, Institute of System Integration in Tsinghua University.

His current research interests include deep reinforcement learning and robot control, especially in continuous control for underwater vehicles.
\end{IEEEbiography}

\begin{IEEEbiography}[{\includegraphics[width=1in,height=1.25in,clip,keepaspectratio]{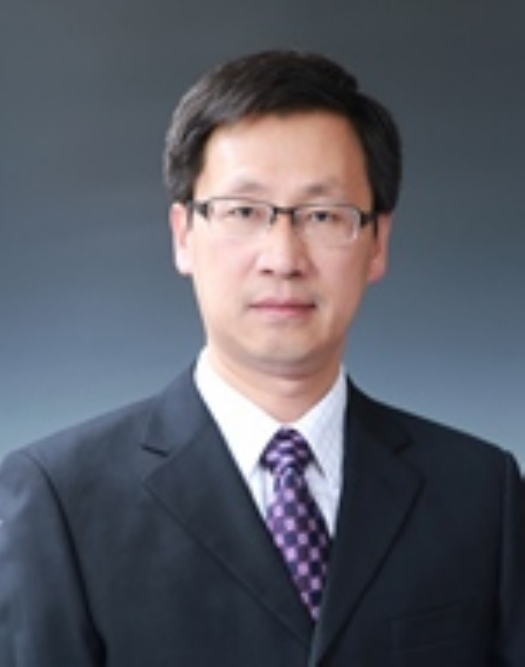}}]{Shiji Song}received the Ph.D. degree in mathematics from the Department of Mathematics, Harbin Institute of Technology, Harbin, China, in 1996.

He is currently a Professor with the Department of Automation, Tsinghua University, Beijing, China. He has authored over 180 research papers. His current research interests include system modeling, optimization and control, computational intelligence, and pattern recognition.
\end{IEEEbiography}

\begin{IEEEbiography}[{\includegraphics[width=1in,height=1.25in,clip,keepaspectratio]{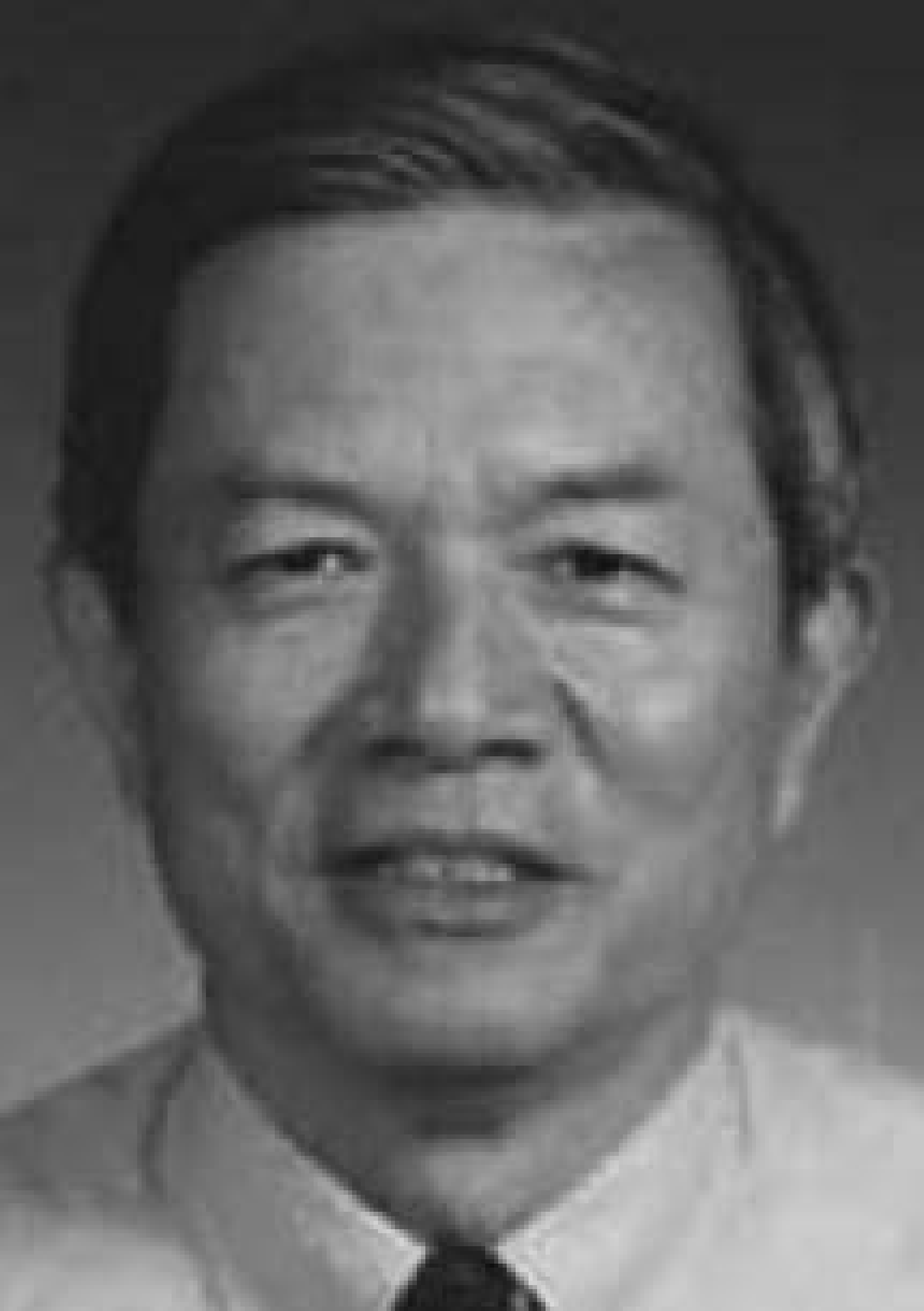}}]{Cheng Wu}received the B.S. and M.S. degrees in electrical engineering from Tsinghua University, Beijing, China.

Since 1967, he has been with Tsinghua University, where he is currently a Professor with the Department of Automation. He has also been an Academician of the Chinese Academy of Engineering since 1995. His current research interests include computer integrated manufacturing systems, manufacturing systems scheduling, optimization of supply chains, and system identification.

Prof. Wu was a recipient of the National Science and Technology Progress Award of China in 1995, 1999, and 2006, and the First Scientific Development Award from the State Educational Committee of China in 1994.
\end{IEEEbiography}

\begin{IEEEbiography}[{\includegraphics[width=1in,height=1.25in,clip,keepaspectratio]{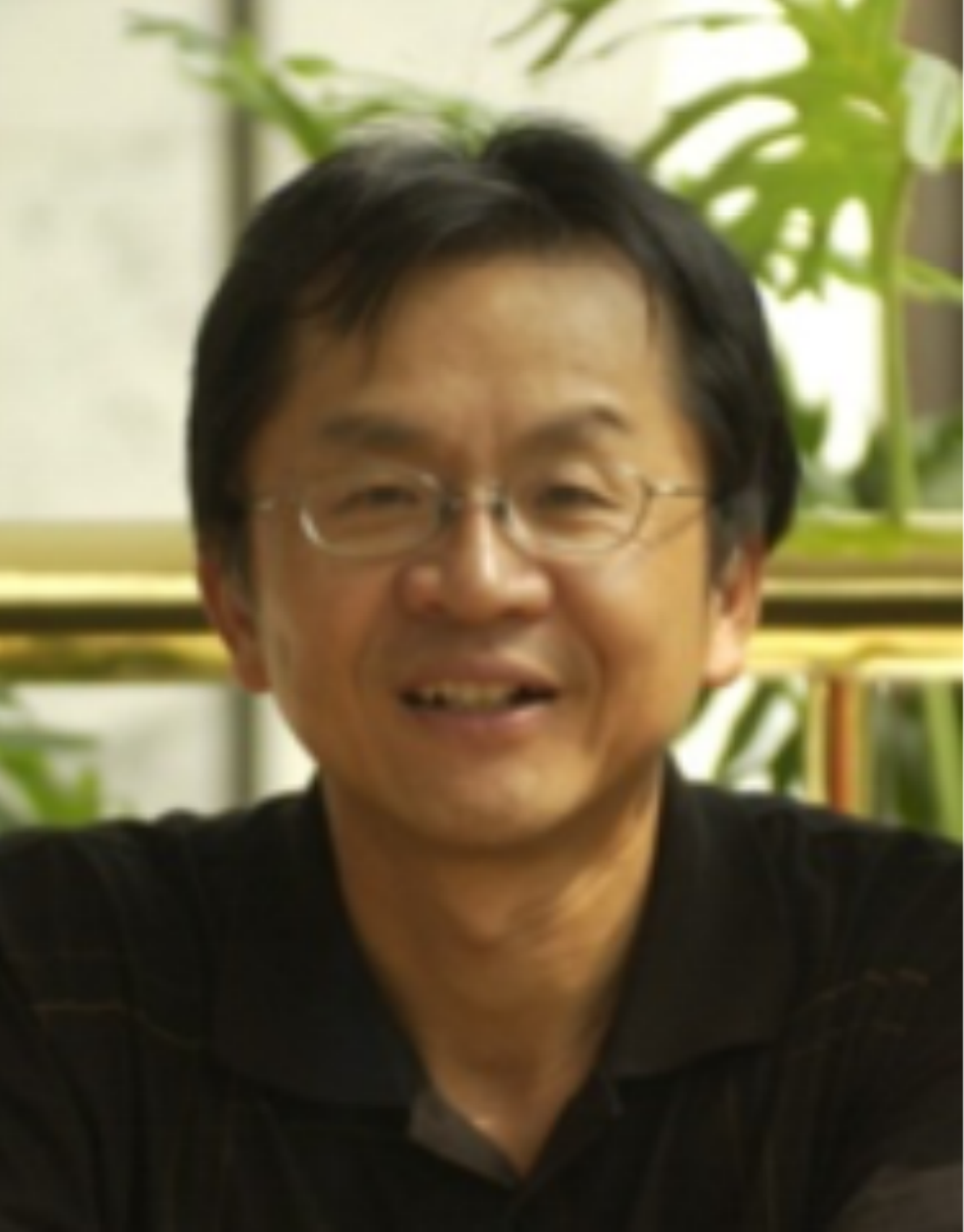}}]{C. L. Philip Chen}received the M.S. degree in electrical engineering from the University of Michigan, Ann Arbor, in 1985 and the Ph.D. degree in electrical engineering from Purdue University, West Lafayette, IN, in 1988.

After having worked at U.S. for 23 years as a tenured professor, as a department head and associate dean in two different universities, he is currently a Chair Professor of the Faculty of Science and Technology, University of Macau, Macau, China.

Dr. Chen is a Fellow of the IEEE and the AAAS. After being the IEEE SMC Society President (2012-2103), currently, he is the Editor-in-Chief of IEEE Transactions on Systems, Man, and Cybernetics: Systems (2014-) and associate editors of several IEEE Transactions. He is also the Chair of TC 9.1 Economic and Business Systems of IFAC. In addition, he is a program evaluator for Accreditation Board of Engineering and Technology Education (ABET) in Computer Engineering, Electrical Engineering, and Software Engineering programs.
\end{IEEEbiography}

\end{document}